\pdfoutput=1
\documentclass[11pt]{article}

\usepackage{acl2023} % USE THIS FOR NON-ANONYMOUS VERSION
 \usepackage{tablefootnote}
\usepackage{wrapfig}
\usepackage{amssymb, amsmath}% http://ctan.org/pkg/amssymb
\usepackage{pifont}% http://ctan.org/pkg/pifont
\newcommand{\xmark}{\ding{55}}%
\usepackage{times}
\usepackage{latexsym}
\usepackage[most]{tcolorbox}

\usepackage[T1]{fontenc}
\usepackage[utf8]{inputenc}
\usepackage{hyperref}
\usepackage{microtype}

\usepackage{inconsolata}

\usepackage{booktabs}
\usepackage{multirow}
\usepackage{newtxmath}
\newcommand{\bspace}{\hspace{1em}}

\title{MATHWELL: Generating Educational Math Word Problems Using Teacher Annotations}

\author{
    Bryan R. Christ\bspace
    Jonathan Kropko\bspace
    Thomas Hartvigsen\\
    University of Virginia\\
    {\small \texttt{brc4cb@virginia.edu,jk8sd@virginia.edu,hartvigsen@virginia.edu}}
}

\begin{document}

\maketitle

\begin{abstract}

Math word problems are critical K-8 educational tools, but writing them is time consuming and requires extensive expertise.
To be educational, problems must be solvable, have accurate answers, and, most importantly, be educationally appropriate.
We propose that language models have potential to support K-8 math education by automatically generating word problems. 
% Existing datasets are unlabeled for these criteria, making them ill-suited for training problem generators. \tom{I removed this because a lack of labels doesn't mean they're bad}
However, evaluating educational appropriateness is hard to quantify. 
We fill this gap by having teachers evaluate problems generated by LLMs, who find existing models and data often fail to be educationally appropriate.
% popular math word problem datasets, namely GSM8K, who find that these datasets substantially fail to be educationally appropriate.
We then explore automatically generating \textit{educational} word problems, ultimately using our expert annotations to finetune a 70B language model.
Our model, MATHWELL, is the first K-8 word problem generator targeted at educational appropriateness.
Further expert studies find MATHWELL generates problems far more solvable, accurate, and appropriate than public models.
% questions 40\% more often than existing open-source models---74\% of generated problems are simultaneously solvable, accurate, and appropriate.
MATHWELL also matches GPT-4's problem quality
% 95\% as good as GPT-4 
 while attaining more appropriate reading levels for K-8 students and avoiding generating harmful questions.\footnote{\href{https://github.com/bryanchrist/MATHWELL}{https://github.com/bryanchrist/MATHWELL}} 

% curating the first teacher-annotated math word problem dataset. %training dataset for this task. 
% We then finetune a 70B language model to serve as a K-8 word problem generator.
% Domain experts find MATHWELL has a 40\% higher share of problems that have executable solutions and meet all criteria than existing open-source models, with 74\% of its problems with executable solutions being solvable, accurate, and appropriate. MATHWELL achieves 94.9\% of GPT-4 Turbo's performance on this task while outputting problems written at a more appropriate reading level for K-8 students and that avoid the most harmful types of inappropriate content. MATHWELL's performance despite being trained by finetuning only highlights the quality of our synthetic data for training appropriate educational word problem generators. We release our experimental code, model, data, and annotations.%\footnote{\href{https://github.com/bryanchrist/MATHWELL}{https://github.com/bryanchrist/MATHWELL}} 
% \footnote{To be made public upon publication.} 
\end{abstract}
\section{Introduction} \label{introduction}
% Paragraph 1: Motivation. What's the problem and why is it important?
% - Focus on the model: Need more test-prep questions, LLMs have been used to generate lots of high-quality data but not for test-prep. There's lots of math LLMs now and so we aim to develop math LLMs to generate test-prep questions.
% Paragraph 2: State-of-the-art and what's missing (most LLMs are answers, not question generators)
% Paragraph 3: Specifically what do we want to do? "We aim to train a question--generator..."
% Paragraph 4: Why is this hard/what's the limitation?
% Paragraph 5: What do we propose in this work and what do we find?
% Paragraph 6: Contributions (bullet-point list, 1 sentence per contribution---go look at examples!)
Math word problems (MWP) are natural language math questions paired with numerical answers and are critical tools for K-12 math education \cite{daroczy_word_2015, article, schwartz_why_2023,verschaffel_word_2020}. Traditionally, teachers hand write MWPs customized to their students' interests, which has been shown to improve students' learning, test performance, and general interest in math \cite{bernacki_role_2018, walkington_using_2013, walkington_personalizing_2019}.
However, teachers' time pressure is often so severe that they must use boilerplate question sets. 
We propose that large language models (LLMs) are poised to enhance math education by generating customized, diverse MWPs for students.
Further, recent advances in math reasoning capabilities in LLMs may imply an approach towards educational impacts.
However, it remains unknown whether math reasoning capabilities \cite{wei_chain--thought_2023} translate to generating \textit{educational} MWPs.

% burdens by automatically generating  exciting opportunity to accelerate question generation, 
% We aim to automatically generate customized math word problems and answers to promote interest-guided math education.

\begin{table*}[t]
    \centering 
    \resizebox{\linewidth}{!}{
    \begin{tabular}{p{.1\linewidth}p{.6\linewidth}p{.3\linewidth}} \toprule
         \textbf{Dataset}& \textbf{Example} &\textbf{Educational Appropriateness} \\ \midrule
         GSM8K&  Henry took 9 pills a day for 14 days. Of these 9 pills, 4 pills cost \$1.50 each, and the other pills each cost \$5.50 more. How much did he spend in total on the pills? \newline
         & \xmark \hspace{.25cm}This question mentions taking pills, which is not appropriate for young learners.\\
          EGSM (Ours)&  Barbie has 100 pink outfits. She has 20 more blue outfits than pink outfits. She has 50\% more green outfits than blue outfits. How many outfits does Barbie have in total?& $\checkmark$ \\ \bottomrule
    \end{tabular}}
    \caption{Example from both GSM8K and EGSM (ours). EGSM is the only teacher validated grade school math dataset, while existing datasets contain problems inappropriate for young learners.}
\label{gsm8k_comparison}
\end{table*}
    \label{tab:example_problem_comparison}
%Systems that generate large volumes of appropriate math problems could accelerate problem customization, improving math education.
% We aim to leverage recent advances in large language models (LLMs) \cite{brown_language_2020} to efficiently automatically generate customized math word problems and answers.
% With the recent proliferation of large language models (LLMs) \cite{brown_language_2020}, it is possible for teachers to use these models to automatically generate customized word problems and answers.
\begin{figure}[t]
    \centering
    \includegraphics[width=0.95\linewidth,angle=270,origin=c]{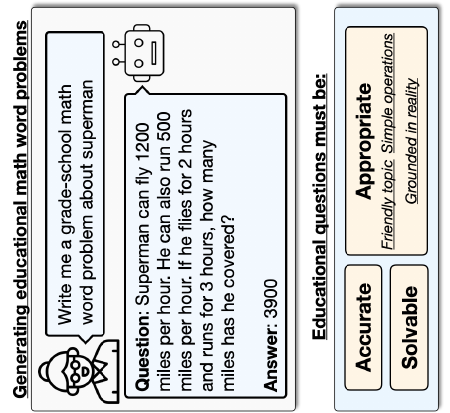}
    \caption{Generating educational math word problems with language models. To be educational, problems must simultaneously be solvable, accurate, and educationally appropriate.}
    % MATHWELLexample generation that meets all evaluation criteria (MaC). See Appendix \ref{sec:prompting} for more details about the prompting process.}
    \label{fig:mathwell_example}
\end{figure}

We aim to explore and enhance LLMs' capacity to generate educational MWPs.
There have been many works using traditional NLP methods to generate MWPs \cite{jiao_automatic_2023, koncel-kedziorski-etal-2016-theme, niyarepola_math_2022, qin_mathematical_2023, qin_math_2024, wang_math_2021, wu_automatic_2022,zhou_towards_2019, zhou2023learning, zong_solving_2023}.
However, they all rely on input \textit{reference} MWPs or equations, ultimately largely rephrasing training data.
To use them, teachers would need to manually curate problem sets, which is time consuming.
And being traditional approaches, these methods are also not promptable, limiting personalization and MWP diversity.
Meanwhile, math reasoning in LLMs has been developing rapidly \cite{wei_chain--thought_2023, yao_tree_2023}.
However, to be useful in education demands strict, domain-specific criteria.
Two recent works explore generating MWPs with LLMs \cite{niyarepola_math_2022,zong_solving_2023}; however, they generate problems without answers.

To evaluate the educational quality of LLMs for math, we recruit real math teachers to assess MWPs generated by SOTA LLMs.
As part of a large human evaluation study, these domain experts spend almost 100 hours evaluating three criteria that make MWPs educational: \textit{Solvability}, \textit{Accuracy}, and, most importantly, \textit{Educational Appropriateness}.
The nuance of appropriateness motivates the need for human evaluations, as it takes years of experience for teachers to develop this sense. Through experiments evaluating existing math reasoning datasets and MWPs generated from five existing LLMs, both public and private models, the teachers identify clear failings in educational appropriateness, especially from open models.%Upon evaluating MWPs from five LLMs, both public and private models, the teachers identify clear failings in educational appropriateness, especially from open models.
  This is somewhat unsurprising, as common math reasoning datasets, like GSM8K \cite{cobbe_training_2021}, are crowd-sourced and not intentionally educational.
However, this may imply we should temper our expectations for direct use of LLMs for elementary math education.

Given a lack of appropriateness in existing datasets' MWPs, we generate the first teacher annotated, educationally appropriate MWP dataset and use it to finetune a new 70B LLM specifically for educational MWP generation.
Experts find that our open model, MATHWELL, matches GPT-4 in educational MWP generation and is 40\% better than the next best open model, generating MWPs that are simultaneously solvable, accurate, and appropriate 74\% of the time.
This performance demonstrates the value of domain expert involvement in developing LLMs for education.
We also find that MATHWELL can be prompted to discuss topics customized to student interests (\textit{e.g.,} Superman) and incorporate specific math operations, outputs MWPs with more appropriate reading levels, and produces fewer harmful errors than other models.
We release our entire human evaluation, including over 5,000 MWPs with gold labels for key educational criteria, along with MATHWELL's training code and weights.

Our key contributions are as follows:
\begin{itemize}
    \item We find existing math datasets are not sufficient to enhance the educational quality of LLM-generated MWPs.
    \item We collect over 5,000 annotations from real teachers, which we filter to create a new, large synthetic training dataset for educational MWP generation, Educational Grade School Math (EGSM).%We collect over 5,000 annotations from real teachers, finding that LLM-generated MWPs largely lack key educational criteria and generating a large synthetic dataset for educational MWP generation.
    \item We use EGSM to finetune a performant LLM for educational MWP generation, which we release alongside our entire human study.
\end{itemize}

\section{Related Work} \label{related_work}
\paragraph{Math QA Datasets} \label{math_qa}
There are many datasets for training and evaluating LLMs on grade school math reasoning. Several popular datasets include GSM8K \cite{cobbe_training_2021}, NumGLUE \cite{mishra_numglue_2022}, GSM-Hard \cite{gao_pal_2023}, ASDIV \cite{miao_diverse_2020} and SVAMP \cite{patel_are_2021}. Recent work has also developed new grade school math evaluation datasets or large training datasets with different solution rationales for existing datasets, particularly GSM8K, including both human written \cite{kim-etal-2023-aint, mishra_lila_2023} and synthetic \cite{mitra2024orcamath, pmlr-v202-shi23a, toshniwal2024openmath, yu2023metamath, yuan_scaling_2023, yue_mammoth_2023} data. 

However, because existing datasets are designed primarily for training and evaluating grade school math reasoning, they are not aligned with training educational grade school MWP generators. Grade school MWP training data for educationally appropriate generators must contain high-quality grammar, be written at an appropriate mathematical difficulty and reading level for K-8 students, include questions similar to those students would encounter in the classroom, and be comprised of questions that are appropriate for an education setting. Further, to encourage effective solution generation, we use Program of Thought (PoT) solutions written as Python functions (see Appendix \ref{important_criteria} for details), as PoT outperforms Chain of Thought (CoT; \citealt{wei_chain--thought_2023}) for open-ended questions, which are the type of problems we seek to generate \cite{azerbayev_llemma_2023, gao_pal_2023, yue_mammoth_2023}. \newline \indent We consider high-quality grammar and similarity to word problems students encounter in the classroom baseline criteria any potentially relevant training data must possess. Existing datasets that contain these baseline characteristics and are aligned with one or more of the other four criteria are GSM-Hard \cite{gao_pal_2023}, GSM8K \cite{cobbe_training_2021}, MathInstruct GSM8K \cite{yue_mammoth_2023}, ASDIV \cite{miao_diverse_2020}, and SVAMP \cite{patel_are_2021}, although none of these datasets contain all four criteria. %For example, while some datasets include PoT solutions (e.g., GSM-Hard and MathInstruct GSM8K), only GSM-Hard contains Python function solutions. Because GSM-Hard contains questions with inflated number values, though, it is not appropriate for training a question generator for K-8 students. 
\paragraph{MWP Generation}\label{word_problem_generation}
Other work explores automatically generating MWPs but requires reference problems or equations as model input and, therefore, constrains the diversity and range of possible outputs and largely rephrases training data. Closest to our work is \citet{zong_solving_2023}, who assess GPT-3's ability to generate MWPs. Their method is reference-dependent, however, as they use a reference problem to guide generation. Further, they only generate MWPs, not solutions. %\citet{niyarepola_math_2022} train LLM word problem generators, but they guide generation with the beginning of a word problem and do not generate solutions. 
\newline \indent Most other works use LLMs or deep neural networks to generate MWPs based on pre-specified equations, provide additional MWPs based on reference problems, or re-write existing MWPS \cite{jiao_automatic_2023, koncel-kedziorski-etal-2016-theme, norberg2023rewriting, niyarepola_math_2022,qin_mathematical_2023, qin_math_2024,wang_math_2021, wu_automatic_2022,zhou_towards_2019, zhou2023learning}, each of which restricts the range of possible outputs. These approaches require additional input from users, which is infeasible for teachers or students who wish to create customized MWPs, making them incomparable to our work. To address this issue, we generate MWP question/answer pairs simultaneously without a reference problem or equation, which we call reference-free generation. To the best of our knowledge, our study is the most comprehensive in exploring the educational appropriateness of generated MWPs alongside teachers. %Further, existing methods have a different use case than reference-free generation, which is more applicable to settings where students need to solve many different types of equations with mixed operations and/or diverse question phrasing, such as end of year assessments or unit tests. As a result, we do not compare MATHWELL against these models/methods. %Our study is the first to evaluate the capability of leading open-source LLMs for the task of MWP generation. 

\section{Methods} \label{methods}
% \paragraph{Our Approach and Evaluation Criteria} \label{approach}
\begin{figure*}[t]
    \centering
    \includegraphics[width=1\linewidth]{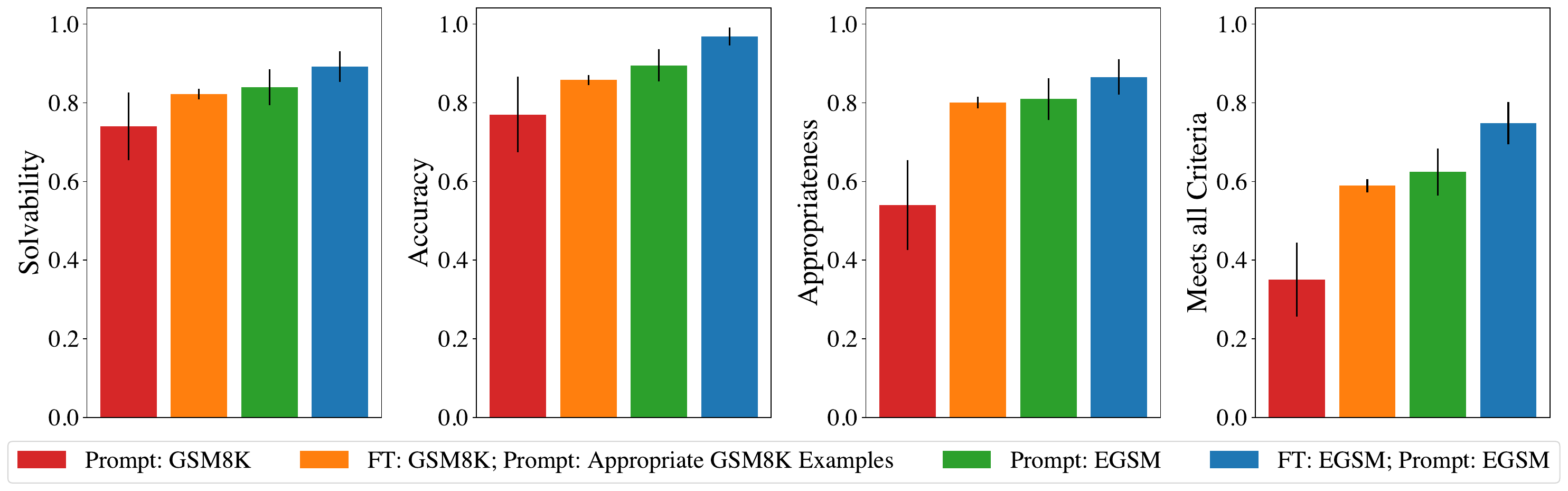}
    \caption{Llama-2 (70B) performance with 95\% confidence intervals on our human evaluation metrics under different prompting/training scenarios. FT is supervised finetuning.}
    \label{fig:barplot}
\end{figure*}
%We propose MATHWELL, a finetuned LLM that generates appropriate word problems aligned with topics K-8 students are interested in and comprised of question/answer pairs.
%MATHWELL can be prompted to generate word problems comprised of question/answer pairs.
%We use MATHWELL to generate a large synthetic dataset designed to train appropriate, educational grade school word problem generators, which we evaluate with human domain experts.
%Figure \ref{fig:mathwell} illustrates our model finetuning and data generation process. \newline \indent 
Math word problems (MWPs) are natural language questions paired with numerical answers.
We study generating these pairs with LLMs, prompting them to write new problems without augmenting hand-picked equations or reference problems. We evaluate these problems based on both the human and automatic evaluation criteria defined below.

\paragraph{Human Evaluation Criteria}
For model-generated MWPs to be educational, they must meet three criteria:
1) \textit{Solvability}, where questions are possible to solve and have one correct answer.
2) \textit{Accuracy}, where generated answers must be correct.
% solution the model generates arrives at the correct answer.
3) \textit{Appropriateness}, where MWPs should be questions teachers would feel comfortable giving to K-8 students.
Generally, appropriate MWPs should make sense, avoid grammatical errors or conflicting information, and be about topics and include mathematical operations appropriate for K-8 students in a school setting.
Because this is hard to define, we emphasize real teacher evaluations of generated MWPs.
% More specifically, educational appropriateness indicates that questions/answers make logical sense, do not have grammatical errors or conflicting information, and are about topics that are appropriate for K-8 students in a school setting.
% Because educational appropriateness is hard to define, we focus on real teacher evaluations of generated MWPs. 
Ideal MWPs are accurate, solvable, and appropriate, thereby meeting all criteria.
We refer to such problems as MaC.
% meet all three criteria, which we denote as MaC.
% We generate questions/answers simultaneously in a reference-free way (e.g., without a reference problem or equation) and with the answer being calculated by a Python function (see Appendix \ref{example_generation} for examples).

%We generate educational MWP question/answer pairs simultaneously in a reference-free way (e.g., without the need for a reference problem or equation). 
\paragraph{Automatic Evaluation Criteria}
Reading level automatically assesses if MWPs are written appropriately. Like \citet{norberg2023rewriting}, we use Flesch-Kincaid Grade Level (FKGL) to evaluate reading level. FKGL is a function of the total words, sentences, and syllables in a text, and the score represents a U.S. grade level \cite{aggarwal_textstat_nodate, flesch_new_1948, kincaid_derivation_1975}. Thus, a FKGL score above 8 for a MWP would be considered inappropriate for an educational K-8 MWP generator. Lower reading levels for MWPs are preferred because high reading levels are known to harm student performance, especially for students who are already struggling \cite{walkington_how_2018}. Negative FKGL scores are possible and denote text that is easy to read due to having short words and sentences. We also calculate each MWP's average token length. Longer token length can be a proxy for MWP complexity, as longer questions tend to include more mathematical operations. 

We follow prior works in using two automatic metrics to compare the quality of our synthetic MWPs to human-written MWPs: Perplexity (PPL) and BERTScore \cite{jiao_automatic_2023,zhou2023learning}. Lower PPL implies better outputs and we calculate PPL using Llama-2 (70B). To show that PPL is not biased towards Llama-2 outputs, we report GPT-2 PPL in Table \ref{tab:additional_auto_eval_models}, finding the same trends in PPL discussed in the sections below. We use BERTScore \cite{zhang2020bertscore} to compute the semantic similarity of our synthetic MWPs and compare it to existing datasets. A lower BERTScore for synthetic MWPs relative to existing datasets would imply they are less similar to each other than human-written MWPs, while a higher score would imply they are more similar. In Section \ref{experiments}, we also use BERTScore to compare our synthetic MWPs to GSM8K to identify if they are similar to human-written MWPs. We calculate GSM8K's within-dataset BERTScore as a reference to compare each source against.

\subsection{Evaluating Existing Datasets} \label{existing_datasets}
We first aim to assess the degree to which existing datasets can be used to prompt models to generate educational K-8 MWPs. We focus our evaluation on GSM8K \cite{cobbe_training_2021} since it is popular and high quality. For each generation, we randomly sample 8 MathInstruct GSM8K samples \cite{yue_mammoth_2023}\footnote{This dataset adds PoT solutions to GSM8K questions.} and use them to few-shot prompt Llama-2 (70B) with a standard prompt asking the model to generate a grade school MWP using PoT to compute numerical answers (prompting details in Appendix \ref{sec:prompting}). We randomly sample 100 generations with executable PoT solutions and acquire teacher annotations for solvability, accuracy, and appropriateness (we discuss further annotation details below). The size of this evaluation sample is consistent with the human evaluation samples used in other MWP generator studies \cite{jiao_automatic_2023, koncel-kedziorski-etal-2016-theme, niyarepola_math_2022, qin_math_2024, wu_automatic_2022,zhou_towards_2019, zhou2023learning, zong_solving_2023}. As shown in Figure \ref{fig:barplot} (red), we find that existing data is ill-suited for prompting models to generate educational word problems, with the worst performance being in appropriateness where barely over 50\% of generations are appropriate, leading to only 35\% of the generations being labeled as MaC. This finding suggests that educationally inappropriate samples such as the one shown in Table \ref{tab:example_problem_comparison} are prevalent in GSM8K. 
%We formulate the answers as Python functions.
% Our generation process takes the following simple notation: $QA = f(p,t)$, where a question/answer pair, \({QA}\), is a function of a standard prompt, \({p}\), and optional topic, \({t}\). See Appendix \ref{sec:prompting} for prompting details.

% Our work has two main outputs: 1) synthetic, model-generated training data and 2) a model that can generate these data.
% Our process for generating these outputs is displayed in Figure \ref{fig:mathwell}. 
\begin{figure}[t]
    \centering
    \includegraphics[width=.42\textwidth]{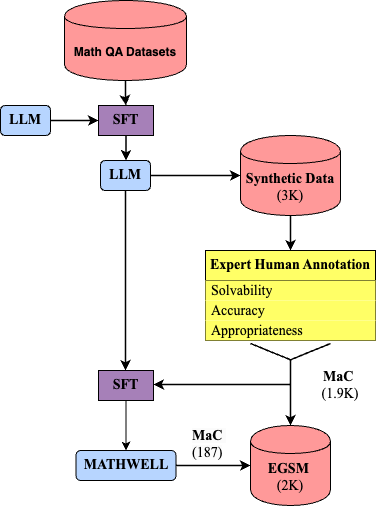}
    \caption{MATHWELL training and EGSM generation process. SFT is supervised finetuning and MaC denotes outputs that meet all criteria.}
    \label{fig:mathwell}
    % \vspace{-20mm}
\end{figure}
\begin{table*}[t]
\centering
\resizebox{\linewidth}{!}{
\begin{tabular}{lcccccccc}
\toprule
\textbf{Dataset}& \textbf{N}& \textbf{PoT}&\textbf{\shortstack{Python\\ Function}}&\textbf{\shortstack{Appropriate\\ Difficulty}} & \textbf{\shortstack{Teacher\\ Annotated}}& \textbf{Length}&\textbf{\shortstack{Reading Level $\downarrow$}}&\textbf{BF1}\\ 
\midrule
 GSM-Hard& 1,319&  $\checkmark$&  $\checkmark$& \xmark  &\xmark& 72.9 (25.6)&4.21 (2.43)&84.0\\
 GSM8K& 8,792& \xmark& \xmark& $\checkmark$  & \xmark& 67.0 (24.4)& 4.24 (2.47)&84.5\\
MathInstruct GSM8K& 6,403&   $\checkmark$&  \xmark&  $\checkmark$  &\xmark& 66.2 (23.9)&4.25 (2.48)&84.6\\
%NumGLUE \cite{mishra_lila_2022}& 12,403&  $\checkmark$ & \xmark & \xmark & 144.8 (136.5)&10.04 (6.99)&10.27 (1.51)&81.5\\
ASDIV& 2,305&  \xmark&  \xmark&  $\checkmark$  &\xmark& 45.1 (15.8)&3.56 (2.40)&85.5\\
 SVAMP& 1,000& \xmark& \xmark& $\checkmark$  &\xmark& 47.3 (11.7)& 3.39 (2.07)&86.1\\
 EGSM (Ours) & 2,093& $\checkmark$& $\checkmark$& $\checkmark$  &$\checkmark$  & 57.2 (15.7)& \textbf{2.50 (1.76)}&85.2\\\bottomrule
\end{tabular}
}
\caption{\label{citation-guide}Characteristics of datasets with more than 1,000 examples that can be used to train reference-free grade school MWP generators. N is the deduplicated number of questions, Length is average question length (in tokens), readability is measured by Flesch-Kincaid grade level, and BF1 is BERTScore F1. Standard deviations, where applicable, are in parentheses.}
\label{dataset_comparison}
\end{table*}

%\subsection{Training MATHWELL} \label{train_mathwell}
% \subsubsection{QLoRA Finetuning} \label{qlora}
%MATHWELL is a finetuned Llama-2 (70B) \cite{touvron2023llama} model that generates question/answer pairs with Program of Thought (PoT) solutions, which recent works have shown improves accuracy \cite{azerbayev_llemma_2023, gao_pal_2023, yue_mammoth_2023}.
%After initial finetuning using public data, namely MathInstruct GSM8K \cite{yue_mammoth_2023}, we prompt our finetuned model to generate synthetic grade school MWPs and acquire expert annotations for solvability, accuracy, and appropriateness. All annotators had K-12 teaching experience and were seasoned educators, with an average of 5.75 years of education experience. They spent an average of 48 seconds per question, totaling 92.33 hours of expert annotation. Further details about the annotation process and annotator agreement are in Appendix \ref{sec:annotation}. Our annotators identified 1,906 problems that are simultaneously solvable, accurate, and appropriate, or meet all criteria (MaC).
%We then finetune MATHWELL further on these MaC outputs to improve performance and demonstrate the value of our synthetic data (see Appendix \ref{sec:finetuning_ablation} for results from each finetuning stage). We conduct all finetuning using QLoRA \cite{dettmers_qlora_2023}. See Appendix \ref{finetuning_details} for more finetuning details.

\subsection{Expert Annotation} \label{EGSM_datasets}
To address the educational inappropriateness of existing math datasets, we generate a high-quality, educationally-appropriate dataset for training MWP generators.
Broadly, we generate synthetic data and evaluate it with teachers (see Figure \ref{fig:mathwell} for our data generation process).
% To create a synthetic data generator with enhanced capability of generating solvable and accurate problems,
To generate this data, we first finetune Llama-2 (70B) \cite{touvron2023llama} using the public MathInstruct GSM8K dataset \cite{yue_mammoth_2023} and QLoRA \cite{dettmers_qlora_2023} (details in Appendix \ref{finetuning_details}). We next identify educationally appropriate GSM8K examples in consultation with teachers and use them to prompt our finetuned model to generate new grade school MWPs.
We then acquire teacher annotations for solvability, accuracy, and appropriateness. All annotators were seasoned educators, with an average of 5.75 years of experience. They spent an average of 42 seconds per question, totaling 55.1 hours of expert annotation. In total, teachers annotated 3,234 synthetic MWPs, with 998 being annotated by two people and 232 annotated by three people. \newline \indent Annotators agreed on solvability \textbf{\(84.6 \pm 2.0\%\) }of the time, accuracy $92.0 \pm 1.5\%$ of the time, appropriateness $74.6 \pm 2.4\%$ of the time, all three labels $66.3 \pm 2.6\%$ of the time, and MaC $76.1 \pm 2.4\%$ of the time. The agreement rates for accuracy and solvability are higher than reported in recent human evaluation studies that analyze human preferences in LLM outputs, and the agreement rates for appropriateness and MaC are on par with these studies \cite{bai_training_2022, ouyang_training_2022, stiennon_learning_2022, ziegler_fine-tuning_2020}. As a result, we feel confident in the quality of our labels. \newline \indent For handling disagreement, if the question was reviewed by two annotators and they disagreed on one of the criteria, we labeled the example as not having the desired criteria. If the question was reviewed by three annotators and there was a disagreement on one of the criteria, we assigned the label with the majority vote. Further details about the annotation process and annotator directions are in Appendix \ref{sec:annotation}. 
\subsection{Further Finetuning on High-quality Outputs}
Based on this annotation process, we identified 1,906 MWPs that are simultaneously solvable, accurate, and appropriate, or meet all criteria (MaC). As shown in Figure \ref{fig:barplot} (orange), the initial finetuning and prompting using educationally appropriate data improves model performance, especially for appropriateness, but still less than 60\% of generations meet all criteria. Therefore, to further improve performance and validate the quality of our synthetic data, we conduct additional finetuning on these MaC outputs to create a model we call MATHWELL. We compile MATHWELL's MaC outputs into a new dataset we call Educational Grade School Math (EGSM).% that combines the samples generated after each finetuning stage.
% To promote future research in training models to generate word problems, we release EGSM,

In total, EGSM contains contains 2,093 MaC question/answer pairs verified by teachers after the first and second stage of finetuning. To support research in automatically scoring model-generated MWPs, we release all our annotated data. To the best of our knowledge, this is the only teacher annotated MWP dataset (see Appendix \ref{sec:annotated_data} for annotated data details). As shown in Figure \ref{fig:barplot} (green and blue), simply few-shot prompting with EGSM outperforms initial finetuning on GSM8K and the additional finetuning on MaC outputs leads to further improvements, showing the second stage of finetuning is critical for educational alignment (see Appendix \ref{sec:finetuning_ablation} for more details on this ablation).

\begin{table*}[t]
\centering
\resizebox{\linewidth}{!}{
\begin{tabular}{llccccccc}
\toprule
& \textbf{Model} & \textbf{Solv.}& \textbf{Acc.}& \textbf{App.}& \textbf{MaC}&\textbf{Top. Spec.}& \textbf{EC }&\textbf{EC/MaC}
\\
\midrule
 % Closed-source Models& & & & & & &\\
 \parbox[t]{2mm}{\multirow{2}{*}{\rotatebox[origin=c]{90}{API}}} &
 GPT-4 Turbo& 94.8 (1.41)& 95.8 (1.31)& 84.4 (2.36)& 78.8 (2.59)& 99.2 (0.56)& 66.8 (2.41)&52.6 (1.73)\\
 & GPT-3.5 Turbo& 88.0 (2.06)& 89.5 (2.07)& 75.5 (2.91)& 62.8 (3.06)& 98.8 (0.69)& 97.5 (0.32)&61.3 (2.99)\\
\midrule
% Open-source Models& & & & & & &\\
\parbox[t]{2mm}{\multirow{4}{*}{\rotatebox[origin=c]{90}{Public}}} &
LLEMMA& 48.8 (3.17)& 63.9 (4.37)&  41.8 (4.48)& 15.2 (2.28)&94.8 (1.41)& 24.3 (0.70)&3.70 (0.55)\\
& MAmmoTH& 86.8 (2.15)& 94.9 (1.49)&  67.7 (3.18)& 56.8 (3.14) &97.6 (0.97)& 6.90 (0.36)&3.91 (0.22)\\
 & Llama-2& 84.0 (2.32)& 89.5 (2.12)& 81.0 (2.72)& 62.4 (3.07)&99.2 (0.56)& 55.4 (0.98)&34.6 (1.70)\\
& MATHWELL (Ours)& \textbf{89.2 (1.97)}& \textbf{96.9 (1.17)}&  \textbf{86.5 (2.29)}& \textbf{74.8* (2.75)}&\textbf{99.6 (0.40)}& \textbf{66.4* (1.00)}&\textbf{49.6* (1.83)}\\
\bottomrule
\end{tabular}}
\caption{\label{citation-guide} Comparing LLMs for MWP generation. All metrics are averages over 250 generations per model for human annotated criteria and over 2,000 for assessing the share of questions with executable code (EC).\tablefootnote{We use a smaller sample size (382) to assess executable code for GPT-4 due to the cost of querying its API.} Solv., Acc., App., MaC, Top. Spec., and EC/MaC are solvability, accuracy, appropriateness, meets all criteria, topic specificity, and the estimated share of questions that MaC and have executable code, respectively. Bold indicates the best open-source performance in each metric and a * indicates the difference between the best open-source performance and second open-source best performance is statistically significant at the p<.01 level. Standard errors are in parentheses.
}
\label{main_results}
\end{table*}
\paragraph{EGSM Dataset Characteristics}\label{EGSM_characteristics}
Table \ref{dataset_comparison} shows the advantages of EGSM over other datasets that have grammatically-correct questions similar to those students encounter in the classroom and are PoT, have Python function solutions, are written at an appropriate difficulty for K-8 students, and/or are educationally appropriate. Critically, EGSM is the only dataset annotated by teachers, ensuring its questions are appropriate for students, which Figure \ref{fig:barplot} shows leads to concrete improvements in human evaluation criteria. Second, EGSM has the lowest average reading level (evaluated by FKGL), so the questions may be more appropriate for those who struggle to read. Third, EGSM is the only dataset with both Program of Thought (PoT) solutions written as Python functions and questions that are mathematically appropriate for K-8 students, which we find critical for training reference-free MWP generators (see Appendix \ref{important_criteria}).\newline \indent EGSM is similar to existing datasets on other common evaluation metrics. While EGSM has a shorter average token length than GSM8K \cite{cobbe_training_2021}, its length is longer than ASDIV \cite{miao_diverse_2020} and SVAMP \cite{patel_are_2021}. This suggests that EGSM's MWP complexity, as reflected in average length, is similar to existing data. EGSM's BERTScore is close to those of existing datasets, suggesting the MWPs are similar. Thus, EGSM is similar to human-written data while being more aligned with educational use.

%(e.g., \citealt{jiao_automatic_2023, niyarepola_math_2022, norberg2023rewriting, zhou_towards_2019, zhou2023learning})

\section{Evaluating MATHWELL} \label{experiments}

\begin{table*}[t]
\centering
\resizebox{\linewidth}{!}{
\begin{tabular}{llcccccccccccccccccc}
\toprule
& & \multicolumn{7}{c}{\textbf{Solvable Questions}} & \multicolumn{7}{c}{\textbf{MaC Questions}}\\
\cmidrule(lr){3-9} \cmidrule(lr){10-16}
& \textbf{Model}& \textbf{Add.}& \textbf{Sub.}& \textbf{Mult.}& \textbf{Div.}& \textbf{Frac.}&\textbf{Dec.}&\textbf{No Ops}& \textbf{Add.}& \textbf{Sub.}& \textbf{Mult.}& \textbf{Div.}& \textbf{Frac.}&\textbf{Dec.}&\textbf{Total Ops}\\
\midrule
% Closed-source Models & & & & & & &&& & & & & &\\
\parbox[t]{2mm}{\multirow{2}{*}{\rotatebox[origin=c]{90}{API}}} &
GPT-4 Turbo& 53.2 &44.7 &61.6 &25.3 &0.80 &4.64 &1.27&54.3&45.2 &59.9 &25.4 &1.02&3.05 & 1.89\\
& GPT-3.5 Turbo& 35.3&28.5 &44.3 & 36.2& 3.17& 23.1&1.81&36.3& 29.9& 39.5& 35.7& 3.18&21.0 &1.66\\
\midrule
% Open-source Models & & & & & & &&& & & & & &\\
\parbox[t]{2mm}{\multirow{4}{*}{\rotatebox[origin=c]{90}{Public}}} &
LLEMMA& 34.4& 27.0& 33.6& 20.5& 6.56& 15.6&15.6&36.8& 39.5& 31.6& 15.8& 2.63& 13.2&1.39\\
& MAmmoTH& 39.6& 37.8& 43.8& 19.4& 3.69& 10.6&2.30& 43.0& 42.2& 40.8& 16.9& 4.93& 9.86&1.58\\
& Llama-2& 57.6& 58.6& 22.9& 14.3& 8.10& 11.4&4.76& 59.6& 60.3& 24.4& 12.8& 5.77&8.97&1.72\\
& MATHWELL (Ours) & 69.5& 69.1& 24.7& 10.3& 5.38& 7.62&1.35& 71.1& 70.6& 24.6& 8.56& 4.81& 7.49&1.87\\
\bottomrule\\
\end{tabular}
}
\caption{\label{citation-guide}Characteristics of model-generated questions. Add., Sub., Mult., Div., Frac., Dec., No Ops, Total Ops, and MaC are addition, subtraction, multiplication, division, fractions, decimals, no operations, total operations, and meets all criteria, respectively. All columns are percentages except total ops, or the average number of distinct operations per question.
}
\label{operations}
\end{table*}
\begin{table}[t]
\centering
\resizebox{\linewidth}{!}{
\begin{tabular}{clccccc}
\toprule
& \textbf{Model}& \textbf{Strange}& \textbf{Too Hard}&\textbf{Harmful}&\textbf{Syntax}&\textbf{No Ops}\\ \midrule
 % Closed-source Models& & & & &\\
\parbox[t]{2mm}{\multirow{2}{*}{\rotatebox[origin=c]{90}{API}}} &
GPT-4 Turbo& 55.3& 13.2& 13.2& 10.5 &7.89\\
& GPT-3.5 Turbo& 60.7& 17.9& 12.5& 1.79 &7.14\\
\midrule
 % Open-source Models& & & & &\\
\parbox[t]{2mm}{\multirow{4}{*}{\rotatebox[origin=c]{90}{Public}}} &
LLEMMA& 41.1& 24.7&6.85&1.37 &26.0\\
& MAmmoTH& 77.9& 7.79&5.19&2.60 &6.49\\
 & Llama-2& 65.1& 4.65&4.65&2.33 &23.3\\
& MATHWELL & 77.4& 0.0&0.0&12.9 &9.68\\
\bottomrule
\end{tabular}
}
\caption{\label{citation-guide}Classification of appropriateness errors for each model. See Figure \ref{fig:appropriateness} for directions presented to annotators and Appendix \ref{sec:examples} for inappropriate samples.
}
\label{tab:appropriateness_classification}
\end{table}

Next, we thoroughly evaluate MATHWELL's generated MWPs using both human and automatic evaluations.
We compare 250 randomly-selected MATHWELL outputs to those of open and closed source models. For closed-source models, we evaluate GPT-3.5 Turbo and GPT-4 Turbo. GPT-3.5 represents a closed-source model with similar capability to Llama-2 (70B) \cite{touvron2023llama}, while GPT-4 is a much more capable model \cite{openai2024gpt4}. For open-source models, we evaluate LLEMMA (34B) \cite{azerbayev_llemma_2023} and MAmmoTH (70B) \cite{yue_mammoth_2023}, which have similar capabilities to Llama-2 (70B) but are math specific.
% We choose to compare MATHWELL to LLEMMA and MAmmoTH because they are two of the leading open-source math-specific LLMs and allow us to determine if existing math LLMs can already succeed at this task \cite{azerbayev_llemma_2023, yue_mammoth_2023}.
%LLEMMA and MAmmoTH are two different approaches to training math LLMs, as MAmmoTH is trained as a QA model \cite{yue_mammoth_2023}, while LLEMMA is trained more generally using a math corpus \cite{azerbayev_llemma_2023}.
Llama-2 serves as a baseline to ensure task-specific finetuning improves performance. 
We prompt each model using examples from EGSM and ask it to create a question/answer pair on a random topic K-8 students are interested in (see Appendix \ref{sec:generation_topics} for a full list of topics). Our 250 evaluation sample for each model is roughly 2.5-5 times larger than the human evaluation samples used in other MWP generator studies \cite{jiao_automatic_2023, koncel-kedziorski-etal-2016-theme, niyarepola_math_2022, qin_math_2024, wu_automatic_2022,zhou_towards_2019, zhou2023learning, zong_solving_2023}.

\subsection{Human Evaluation} \label{human_eval}
\paragraph{MATHWELL Matches SOTA in Human Evaluation Criteria}%\label{sec:main_results}
Teachers scored the 250 MWPs from each model for solvability, accuracy, and appropriateness (see examples in Appendix \ref{sec:examples} and additional annotation details in Appendix \ref{sec:annotation}).
% Appendix \ref{sec:examples} contains samples for each model that do and do not meet each criteria, along with a discussion on why the sample does not meet the specified criteria.
Annotators also assessed topic specificity, or if the MWP incorporates the specified topic. Table \ref{main_results} shows MATHWELL performs best among open-source models in all metrics, with a 19.9\% higher share of outputs that MaC, 19.9\% higher share that have executable code, and 43.4\% higher share that have executable code and MaC than the next best open-source model. MATHWELL also outperforms GPT-3.5 in human evaluation criteria and matches 94.9\% of GPT-4's performance in MaC, outperforming it in accuracy and appropriateness. 

Although MATHWELL performs best in topic specificity, the other models do well in this metric, implying that LLMs can effectively customize MWPs to student interests. Table \ref{main_results} also highlights that the open-source models lag significantly behind GPT-4 in human evaluation criteria, but that QLoRA finetuning on EGSM is enough to match its performance. Lastly, Table \ref{main_results} highlights that open-source models have a large gap in the share of question/answer pairs that contain executable code relative to GPT-3.5, the best performing model in this metric (see Appendix \ref{sec:gpt4_coding} for details on GPT-4's performance on this metric). %Specifically, the math models struggle with appropriateness and generating executable code, showing that raw coding ability does not necessarily translate into more outputs containing executable code. 
%Table \ref{main_results} also suggests that QLoRA finetuning on our high-quality data is enough to match SOTA models' performance on human evaluation criteria for this task. 
\paragraph{MATHWELL Generates High-quality, Complex Questions} \label{question_complexity}
MWPs involving multiplication, division, fractions, and decimals are more complex than those involving addition and subtraction, and we aim to train a generator capable of creating MWPs from each of these operations K-8 students regularly encounter. In Table \ref{operations}, we assess whether each model can generate MaC questions when using more complicated operations. Questions may contain more than one operation. Solvable questions may require no math operation if they contain the answer in the question (see Appendix \ref{no_operation_example} for an example), so we also report the share of questions containing no operations. To determine if questions with complex operations are accurate and appropriate, we compare the math operations in solvable questions to those in MaC questions. We also report the average number of distinct operations in MaC questions for each model, which is another way to assess question complexity.
\newline \indent Table \ref{operations} shows that GPT-4 and MATHWELL are the only models for which the share of MaC questions for each operation is within two percentage points of that for solvable questions, showing MATHWELL can generate MaC questions regardless operation complexity. GPT-4 and MATHWELL are also the least likely to generate problems that require no operations and have the highest average total operations, which also suggests MATHWELL generates high-quality, complex problems. These findings suggest MATHWELL can effectively generate MWPs that cover the full range of problem types K-8 students encounter.
\newline\indent We do not prompt models for specific operations. Under these conditions, MATHWELL generates more problems containing addition and subtraction relative to the other models. In turn, there is a concern that MATHWELL’s performance in Table \ref{main_results} could be due to it generating simple questions for this experiment, which may be more likely to MaC. To address this concern, we conduct two additional analyses reported in Appendix \ref{sec:additional_analyses}: 1) logistic regressions showing MATHWELL's higher MaC relative to the other models except GPT-4 remains statistically significant when controlling for math operations and 2) a summary of accuracy by operation showing GPT-4 and MATHWELL are the only models for which accuracy does not substantially differ by operation and remains above 90\% for each operation. These results buttress our finding that MATHWELL can generate MaC MWPs regardless of operation and pinpoint the operations for which each comparison model most commonly fail. 

\paragraph{MATHWELL Makes Less Severe Appropriateness Errors} \label{appropriateness_errors}
As shown in the annotator directions in Figure \ref{fig:appropriateness}, there are four primary reasons why a question would be flagged as inappropriate: being strange or unrealistic,
% (see Appendix \ref{strange_question} and \ref{unrealistic_question} for examples), 
being too difficult for a K-8 student, 
% (see Appendix \ref{too_hard_example} for an example), 
containing inappropriate content for a classroom setting, 
% (see Appendix \ref{inappropriate_content} for an example),
or having grammatical errors or typos. A final reason why a question would be labeled as inappropriate is that it does not require any mathematical operations to solve and, therefore, is a reading comprehension question rather than a MWP. 
% (see Appendix \ref{no_operation_example} for an example).
See Appendix \ref{sec:examples} for examples of each.
Of these errors, questions that contain inappropriate content or are too difficult are the most harmful for young learners. Questions that are strange/unrealistic, have typos, or assess reading comprehension rather than math reasoning are not good, but they do not present a harm to student learning. As shown in Table \ref{tab:appropriateness_classification}, MATHWELL is the only model that does not make these more harmful errors, showing that it is better suited for generating MWPs that are educationally appropriate for young learners. 

\subsection{Automatic Evaluation} \label{auto_eval}
\begin{figure}[t]
    \centering
    \includegraphics[width=1\linewidth]{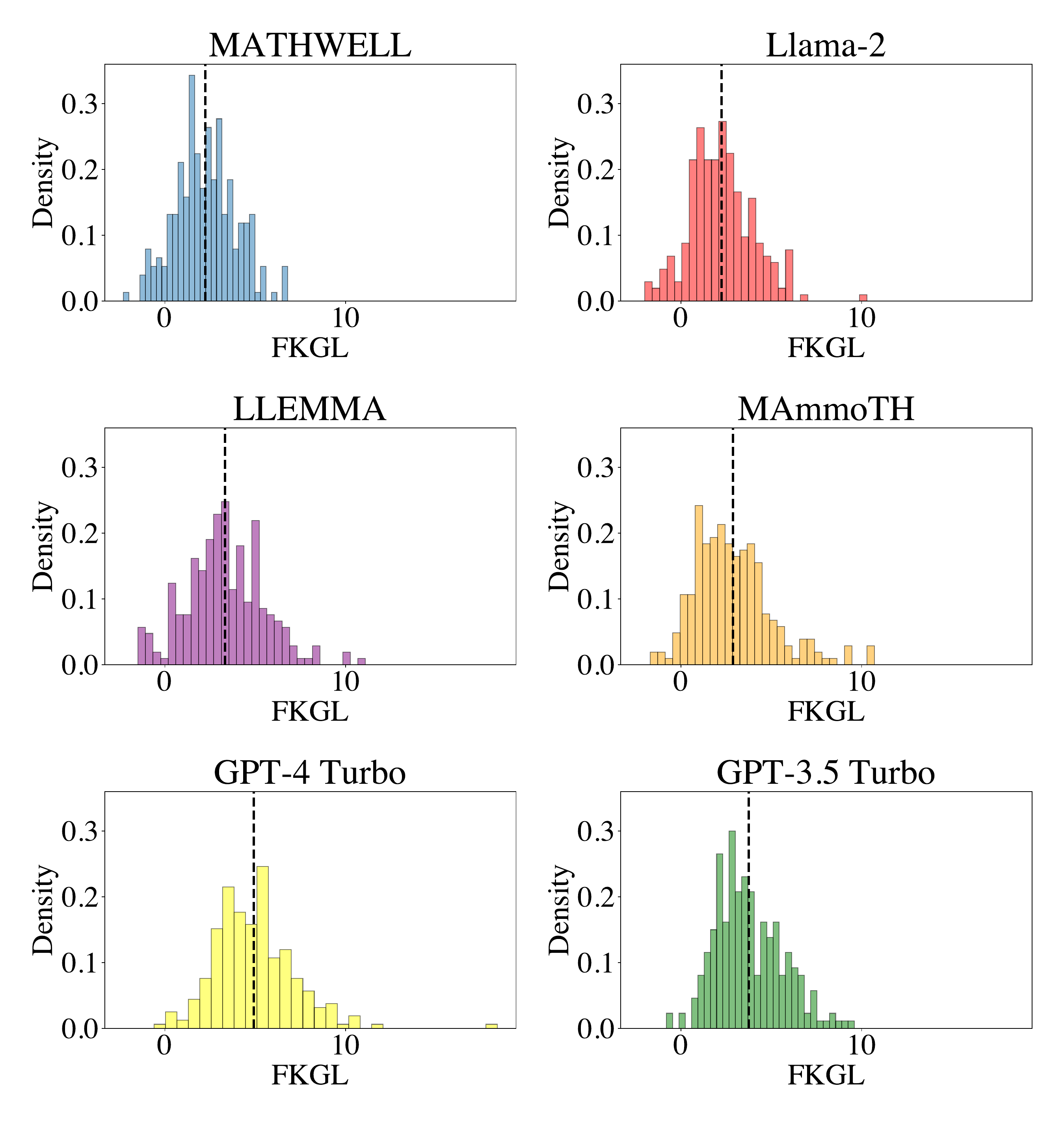}
    \caption{Flesch-Kincaid grade level (FKGL) distribution of model MWPs. Dotted lines show mean FKGL.}
    \label{model_read}
\end{figure}
\paragraph{MATHWELL Outputs Have More Appropriate Readability} \label{flesch}
Figure \ref{model_read} compares the FKGL distribution of MATHWELL outputs to those of the other models considered. As shown in the figure, MATHWELL has the lowest FKGL and is the only model that does not generate questions beyond an 8th grade reading level, providing evidence it creates more readable questions for K-8 students. In Figure \ref{dataset_read}, we see a similar trend in readability for EGSM: Fewer of EGSM's MWPs than other existing datasets have a FKGL score greater than 8.

\paragraph{MATHWELL Outputs Are Human Quality}\label{sec:ppl}
Table \ref{tab:auto_eval_datasets} shows EGSM has lower PPL than GSM8K and Table \ref{tab:auto_eval_models} shows MATHWELL's 250 evaluation sample for the experiments in Table \ref{main_results} has lower PPL than that of the other models. As shown in Tables \ref{tab:auto_eval_datasets} and \ref{tab:auto_eval_models}, across models, datasets and comparisons, BERTScores are similar. Taken together, these findings suggest our synthetic data are similar to human quality, so our MWPs will likely be similar to ones students normally encounter.

\begin{table}[t]
\centering
\resizebox{1\linewidth}{!}{
\begin{tabular}{lcc}
\toprule
\textbf{Dataset}& \textbf{PPL $\downarrow$} & \textbf{BERTScore F1}\\
\midrule
 GSM8K& 4.05 (1.18)& 84.5\\
 EGSM& 2.44 (0.439)& 84.3\\ \bottomrule
\end{tabular}
}
\caption{\label{citation-guide}Automatic evaluation for datasets. PPL is perplexity and BERTScore F1 compares each dataset's questions to GSM8K.
%, and GSM8K BF1 compares each dataset's questions to MathInstruct GSM8K. 
Bold denotes the lowest PPL. Standard deviations, if applicable, are in parentheses.
}
\label{tab:auto_eval_datasets}
\end{table}

\begin{table}[t]
\centering
\resizebox{\linewidth}{!}{
\begin{tabular}{clcccc}
\toprule
& \textbf{Model}& \textbf{PPL $\downarrow$} & \textbf{BF1}&\textbf{GSM BF1}&\textbf{MaC Length}\\ \midrule
 % Closed-source Models& & & & &\\
\parbox[t]{2mm}{\multirow{2}{*}{\rotatebox[origin=c]{90}{API}}} &
GPT-4 Turbo& 2.50 (0.03)& 85.4& 84.6& 66.8 (2.62)\\
& GPT-3.5 Turbo& 2.64 (0.03)& 85.6& 84.6& 49.9 (1.16)\\
\midrule
 % Open-source Models& & & & &\\
\parbox[t]{2mm}{\multirow{4}{*}{\rotatebox[origin=c]{90}{Public}}} &
LLEMMA& 3.82 (0.10)& 84.3&84.0&50.9 (2.89)\\
& MAmmoTH& 2.76 (0.03)& 86.0&84.6&44.4 (1.15)\\
 & Llama-2& 2.52 (0.03)& 85.5&84.3&49.8 (1.19)\\
& MATHWELL& \textbf{2.44 (0.03)}& 85.5&84.2&\textbf{54.1 (0.97)}\\
\bottomrule
\end{tabular}
}
\caption{\label{citation-guide}Automatic evaluation metrics for each model. PPL is perplexity, BF1 is BERTScore F1, GSM BF1 compares each model's questions to GSM8K, MaC is meets all criteria, and Length is average token length. Bold indicates the lowest PPL and longest length for MaC generations for open-source models. Standard errors, where applicable, are in parentheses.
}
\label{tab:auto_eval_models}
\end{table}

\paragraph{MATHWELL Outputs Longer MaC Questions than Existing Open-source Models}\label{sec:question_length}
Longer token length may signal more complex MWPs, as longer MWPs often contain more information and operations than shorter ones. Comparing the average length of all MWPs to the length of MaC MWPs can determine if MaC MWPs are shorter or simpler. As shown in Table \ref{tab:auto_eval_models}, MATHWELL's MaC MWPs are the longest of the open-source models considered, suggesting its MaC problems may be more complex. As shown in Table \ref{tab:additional_auto_eval_models}, MATHWELL and GPT-4 are also the only models for which MaC length is within a token of the overall average, providing evidence that its MaC MWPs are no simpler than its average MWP. While GPT-4 outputs longer MWPs than MATHWELL on average, MATHWELL’s MWP length is still on par with those of the human-written grade school math datasets reported in Table \ref{dataset_comparison}. This finding further supports those discussed above showing that MATHWELL generates MWPs from a range of difficulties and complexities, which implies coverage of concepts throughout K-8 math education.

\section{Controllability} \label{controllability_experiment}\begin{table}[t]
\centering
\resizebox{\linewidth}{!}{
\begin{tabular}{lcccc}
\toprule
& \textbf{Add.}& \textbf{Sub.}& \textbf{Mult.}& \textbf{Div.}\\
\midrule
% Closed-source Models & & & & & & &&& & & & & &\\
Prompted for Operation (n=400)& \textbf{64.0} &\textbf{77.0}&\textbf{67.0} &\textbf{44.0}\\
Unprompted& 6.07&10.12&6.07&1.62\\
\bottomrule
\end{tabular}}
\caption{\label{controllability_results}Percentage of questions generated that include only one operator when prompted to do so, compared to overall percentages that contain only that operator. Add., Sub., Mult., and Div. are addition, subtraction, multiplication, and division, respectively.}
\end{table}
Being able to control the math operations and topics present in MWPs generated by MATHWELL would maximize the model's educational applicability by allowing teachers to generate MWPs that target specific concepts they are teaching. To test MATHWELL's controllability, we prompted the model to generate questions involving only one of the operations from the list of Addition, Subtraction, Multiplication, and Division. Each prompt included 8-shot in-context examples and we generated 100 questions for each operation, totaling 400 generated questions. See Appendix \ref{sec:prompting} for the full prompt. We then measured the percent of generated questions that successfully include only the intended operator. We compare this to the distribution of questions produced by the unprompted model that only contain one of the operators.

As shown in Table \ref{controllability_results}, we find that requesting specific operators significantly increases the rate at which they are generated. There is room for improvement, especially for division, which we believe is clear future work. Regardless, this rate of operation controllability is competitive with the rate at which existing SOTA word problem generators successfully incorporate a pre-specified equation, with the exception of division controllability, which is similar to the rates of controllability reported in many of these works \cite{qin_mathematical_2023, qin_math_2024, wang_math_2021, wu_automatic_2022, zhou_towards_2019, zhou2023learning}. These results paired with the high rate of topic controllability reported in Table \ref{main_results} demonstrate that our model is non-trivially controllable.
\section{Conclusions} \label{conclusions}
We explore educational reference-free K-8 MWP generation and create the first dataset to train models for this task and the only one with teacher annotations. We demonstrate the quality of this dataset by using it to finetune MATHWELL, the first reference-free educational MWP generator. Our evaluations show that MATHWELL outperforms other open-source LLMs at reference-free MWP generation, matching the performance of GPT-4 while generating questions with a more appropriate reading level and that do not contain harmful inappropriate content. Further, we show that our dataset, EGSM, is comparable to existing math QA data while containing questions that are more educationally appropriate. These findings suggest that reference-free MWP generation is a feasible and practical alternative to traditional reference-dependent generators. Future research should train reference-free MWP generators to create questions aligned with specific grade levels and develop automated classification methods to reduce human annotation costs. 

%MATHWELL, which generates a question/answer pair based only on an optional topic. To train our model, we generate synthetic data and use expert annotators to identify a high-quality training subset. We release EGSM, a synthetic dataset of 20,490 question/answer pairs for use in future research. Our evaluations show that MATHWELL outperforms other open-source LLMs at reference-free word problem generation, nearly matching the performance of GPT-4 on this task while generating questions with a more appropriate reading level, and that EGSM is of comparable quality to existing math QA data. These findings suggest that reference-free word problem generation is a feasible and practical alternative to traditional context-dependent generators. Future research should train reference-free word problem generators that can create questions aligned with specific math topics and grade levels and further develop automated classification methods to reduce human annotation costs. 

\section*{Limitations} \label{limitations}
One limitation of MATHWELL is that it is not specifically designed to generate questions aligned with pre-specified grade levels, which we chose to leave to future research due to the high cost of annotating questions for these characteristics. However, we demonstrate that our model is non-trivially promptable for specific math operations in Section \ref{controllability_experiment}, which highlights that exploring additional controllability is a promising area for future research. Additionally, MATHWELL is trained and evaluated for generating MWPs/solutions for K-8 students only; therefore, we do not recommend using it to generate question/answer pairs for other grade levels or for other school subjects, which we believe are compelling areas for future work. Another limitation of MATHWELL's MWPs is that they are text-only, and students often encounter MWPs that are multi-modal (containing both images/tables/figures and text) in addition to those that are text-only. Generating multi-modal problems is thus a challenging and interesting area for future work. While we use a standard prompt for our experiments to simplify the evaluation framework and make fair comparisons across models, future work could conduct prompt tuning experiments to further improve model performance. 

The high cost of human evaluation to ensure educational outputs is another limitation of automatic MWP generation broadly. In Appendix \ref{sec:classifier}, we explore training text classifiers to automatically score MATHWELL outputs. We show existing models can learn some features important for automatic classification, but need refinement in order to correctly classify unsolvable, inaccurate, or inappropriate questions. We hope these results and our large annotated dataset motivate future research in automatic classification efforts. 
\newline \indent Another limitation of this work is the subjective nature of the appropriateness criteria. While it is critical model-generated questions are appropriate for students, it is hard to fully define all aspects of appropriateness and individuals may have differing opinions on the degree to which a question is appropriate or not. We chose to define several common reasons questions may not be appropriate for students (see Figure \ref{fig:appropriateness}) and use teachers as annotators, but future research should continue to define this criteria and include multiple evaluators.

\section*{Ethics Statement}\label{ethics}
All data used to train MATHWELL come from open-access datasets and, therefore, should not contain any private sensitive information. MATHWELL may generate questions that are inappropriate for use in educational contexts and additional research should be conducted on the model before deploying it in classroom settings. Specifically, future research should continue to improve performance of text classifiers to filter out questions which are not appropriate for students.

\section*{Acknowledgements} \label{acknowledgements}
We thank Zooniverse \cite{noauthor_zooniverse_nodate} for providing a free and user-friendly platform for data annotation. We also thank our expert volunteer annotators for providing high-quality labels and feedback on our evaluation criteria/directions. 
Finally, we thank the University of Virginia's Research Computing team for maintaining excellent high-performance computing resources that allowed us to conduct this research.

\bibliography{anthology,custom}
\bibliographystyle{acl_natbib}

\appendix
\section{Prompting Process}\label{sec:prompting}
\subsection{Standard Prompts and Sampling Process}
For all generations reported in this paper, we set temperature to 1.0 to strike a balance between creativity and encouraging the model to select the most probable next token to guide effective solution generation. We also set the sampling parameter equal to true to further diversify outputs while also ensuring the most probable next token still receives the most weight. The GitHub repo associated with this paper has a sample generation script that uses this sampling approach. The exact prompts used in each experiment are reported below.

\paragraph{Prompt for Generating Synthetic Training Data}
Our standard prompt for evaluating the educational appropriateness of existing datasets (Section 
\ref{existing_datasets}) and generating synthetic training data after the first stage of finetuning on public data (Section \ref{EGSM_datasets}) is, "Write a grade school math word problem and Python function with a commented out step-by-step solution to solve the word problem." We begin each prompt with 8-shot examples, using randomly selected examples from MathInstruct GSM8K for evaluating existing datasets and a consistent set of 8 educationally appropriate MathInstruct GSM8K examples for generating synthetic training data. 

\paragraph{Prompt for Interacting with MATHWELL and Comparing Models}
Our standard prompt for interacting with the MATHWELL model and for all experiments reported in Section \ref{experiments} is, "Write a grade school math word problem about \{topic\} and Python function with a commented out step-by-step solution to solve the word problem." 

In this prompt, topic is an optional argument, which we randomly select from the list of topics discussed in Appendix \ref{sec:generation_topics}. We begin every prompt with a random selection of 8-shot examples from EGSM.

\paragraph{Prompt for Controllability Experiment:}
Our prompt for the controllability experiment reported in Section \ref{controllability_experiment} is: "Write a grade school math \{operation\} word problem about \{topic\} and Python function with a commented out step-by-step solution to solve the word problem. The question you write should only require \{operation\} to solve, meaning the solution should rely only on use of the \{operator\} operator." 

In this prompt, operation is filled in with addition, subtraction, multiplication, or division, topic is randomly selected from the list of topics in Appendix \ref{sec:generation_topics}, and operator is filled in with +, -, *, or /. We begin every prompt with a random selection of 8-shot examples from EGSM that contain only the desired operation.

\subsection{Expected Output Format}\label{example_generation}
Using the prompting processes described above, the prompted model should output a question/answer pair with the solution being calculated by a Python function. The response should separate the question and solution using "Question:" and "Solution:," respectively. Below is a sample output from MATHWELL showing this output structure. \newline 

\noindent \textbf{Sample Generation:}\newline
\noindent Question: Superman can fly 1200 miles per hour. He can also run 500 miles per hour. If he flies for 2 hours and runs for 3 hours, how many miles has he covered?\newline \newline
Solution: \newline
def solution():\newline
\indent \#Superman can fly 1200 miles per hour\newline
\indent superman\_flight\_speed = 1200 \newline
\indent \#He can run 500 miles per hour \newline
\indent superman\_run\_speed = 500 \newline
\indent \#He flies for 2 hours \newline
\indent flight\_hours = 2 \newline 
\indent \#He runs for 3 hours \newline 
\indent run\_hours = 3 \newline
\indent \#The answer is\newline
\indent result = (superman\_flight\_speed * flight\_hours) \newline \indent \indent + (superman\_run\_speed * run\_hours) \newline
\indent return result

\subsection{Additional Suggested Prompting Strategies}
We find MATHWELL is more likely to generate executable code when given a topic than when a topic is not specified. For example, when prompting our finetuned Llama-2 model before further training it on the EGSM data, we found the model generated executable code 63.1\% of the time when given a topic, and only 32.3\% of the time when a topic was not specified. As a result, for evaluating models in this paper, we provide them with a randomly selected topic, which also gives us the ability to assess their ability to effectively generate topic-specific word problems. Additionally, this evaluation strategy is aligned with how a teacher or student would use the model in practice, as they would want the generated questions to align with a particular topic. 
Qualitative evaluations of model generations also revealed that MATHWELL is more likely to generate executable code when the topic is more specific. For example, if their desired topic is superheroes, a user would have a higher likelihood of receiving a generation with executable code by prompting with a specific superhero (e.g., Superman) than leaving the topic general (e.g., superheroes).

\section{Annotation Process and Details}\label{sec:annotation}
\begin{figure}[t]
    \centering
    \includegraphics[width=1\linewidth]{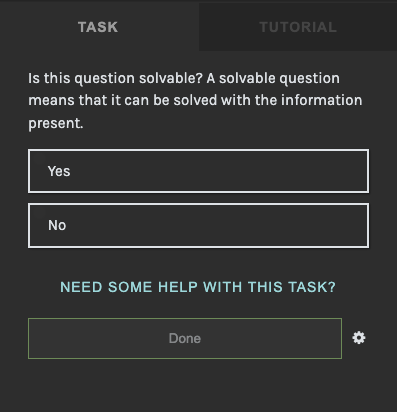}
    \caption{Solvability directions.}
    \label{fig:solvability}
\end{figure}
\subsection{Annotators}
All annotators had K-12 teaching experience or training, including a research team member who annotated every question. We had three primary annotators who reviewed at least 200 questions each in addition to our research team member. Our four primary annotators were seasoned educators, with an average of 5.75 years of education experience. They spent an average of 48 seconds per question, totaling 96.3 hours of expert annotation throughout our full human evaluation of over 5,000 model-generated word problems.
\begin{figure}[t]
    \centering
    \includegraphics[width=1\linewidth]{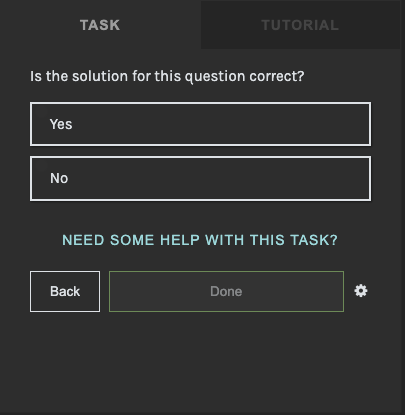}
    \caption{Accuracy directions.}
    \label{fig:accuracy}
\end{figure}
\begin{figure}[t]
    \centering
    \includegraphics[width=1\linewidth]{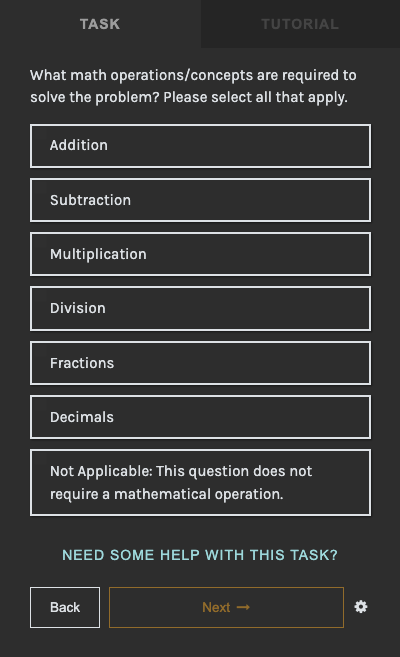}
    \caption{Labeling operations directions.}
    \label{fig:operations_directions}
\end{figure}

\subsection{Validating Final Evaluation Labels}
For annotating the 250 samples from each model for our experiments reported in Section \ref{experiments}, we randomized questions from each model and had them blindly reviewed by one of our highly trained annotators with K-12 teaching experience. To evaluate the quality of these labels, we had 357 randomly reviewed by one additional annotator and 60 randomly reviewed by two additional annotators. The annotators agreed on solvability \textbf{\(89.3 \pm 3.0\%\) }of the time, accuracy $95.5 \pm 2.0\%$ of the time, appropriateness $80.4 \pm 3.8\%$ of the time, all three labels $67.8 \pm 4.5\%$ of the time, and MaC $78.3 \pm 3.9\%$ of the time. The agreement rates for accuracy and solvability are higher than reported in recent studies that explore human alignment of LLM outputs, and the agreement rates for appropriateness and MaC are on par with these studies \cite{bai_training_2022, ouyang_training_2022, stiennon_learning_2022, ziegler_fine-tuning_2020}. 
\begin{figure}[t]
    \centering
    \includegraphics[width=1\linewidth]{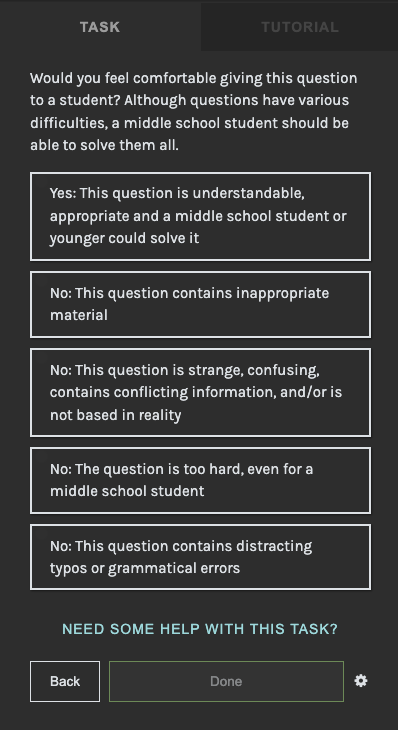}
    \caption{Appropriateness directions.}
    \label{fig:appropriateness}
\end{figure}
Additional analysis reveals that most annotator disagreement (80.1\%) was due to the primary annotator being more conservative than the additional annotators by labeling questions as not having the desired criteria when the additional annotators rated them as having the desired criteria. As a result, we chose to use the labels from the primary annotator when reporting final results to be conservative, though we also found the results do not vary when switching labels based on annotator disagreement. Annotators were least likely to disagree on labels for MATHWELL and GPT-4 Turbo outputs and our primary annotator was not more likely to rate MATHWELL outputs as having the desired criteria than the additional annotators. Taken together, this evidence suggests our final labels are highly accurate. 

\subsection{Defining Appropriateness}
Teachers participated in designing our appropriateness metric. Our research team includes a former K-12 teacher, who led metric development. We iteratively refined the metric with feedback from teacher annotators using sample model outputs. This process led to the definition described in the paper along with the annotator directions and response options shown in Figure \ref{fig:appropriateness}.

\subsection{Annotation Interface}
We used Zooniverse \cite{noauthor_zooniverse_nodate} to collect our human annotation data. Figures \ref{fig:solvability}, \ref{fig:accuracy}, \ref{fig:operations_directions} and \ref{fig:appropriateness} show the instructions each annotator was given for each of our evaluation criteria. 

\subsection{Annotated Data Characteristics}\label{sec:annotated_data}
Our full annotated dataset consists of 5,084 question/answer pairs with teacher annotations for solvability, accuracy, appropriateness, and MaC. The data are comprised of the 3,234 word problem dataset used to generate training data for MATHWELL in addition to the 250 evaluation set for each model described in Section \ref{experiments}, the 100 questions we evaluated in Section \ref{existing_datasets}, and 250 questions we evaluated in Appendix \ref{sec:finetuning_ablation} from training our model using the 2nd finetuning stage only. Based on our annotations, 82.2\% of the question/answer pairs are solvable, 87.3\% have accurate solutions, 78.1\% are appropriate, and 58.4\% meet all criteria. 

\section{Additional Experiments}\label{sec:additional_analyses}
\subsection{Finetuning Ablation}\label{sec:finetuning_ablation}
\begin{table*}[t]
\centering
\resizebox{\linewidth}{!}{
\begin{tabular}{lcccccc}
\toprule
\textbf{Model}& \textbf{Solv.}& \textbf{Acc.}& \textbf{App.}& \textbf{MaC}& \textbf{EC }&\textbf{EC/MaC}
\\
\midrule
 MATHWELL (1st Stage Only)& 82.3 (0.67)& 85.8 (0.66)& 80.1 (0.77)& 58.9 (0.87)& 63.1 (0.73)&37.2 (0.55)\\
 MATHWELL (2nd Stage Only)& 86.0 (2.20)& 91.6 (1.89)& 76.7 (2.89)& 60.8 (3.09)& 50.9 (1.20)&31.0 (1.58)\\
MATHWELL (1st and 2nd Stage)& \textbf{89.2 (1.97)}& \textbf{96.9* (1.17)}&  \textbf{86.5* (2.29)}& \textbf{74.8** (2.75)}& \textbf{66.4** (1.00)}&\textbf{49.6** (1.83)}\\
\bottomrule
\end{tabular}}
\caption{\label{citation-guide}Average metrics for MATHWELL after the first and second round of finetuning compared to the model finetuned with both stages. Solv., Acc., App., MaC, and EC/MaC are solvability, accuracy, appropriateness, meets all criteria, and the estimated share of questions that MaC and have executable code, respectively. Bold indicates the best performance in each metric and a * or ** indicates the difference between the best performance and second best performance is statistically significant at the p<.05 or p<.01 level, respectively. Standard errors are in parentheses.
}
\label{tab:finetuning_ablation}
\end{table*}
As shown in Table \ref{tab:finetuning_ablation}, MATHWELL shows a statistically and/or substantively meaningful improvement in all metrics after the second stage of finetuning on high-quality synthetic data labeled by domain experts relative to just doing either finetuning stage alone. These results lead us to conclude that the second stage of finetuning is critical for improving the model's ability to generate questions that are educationally appropriate, or MaC. 

\subsection{Logistic Regression for Predicting MaC}
As shown in Table \ref{mac_regression}, the coefficients for all models except GPT-4 for MaC remain negative and statistically significant relative to MATHWELL when controlling for the type of mathematical operation. This finding supports the assertion that MATHWELL is more capable than these models of generating MaC questions regardless of the operation considered, even if it is less likely to generate questions from the more complex mathematical operations. Table \ref{mac_regression} also shows that the difference in MaC performance between MATHWELL and GPT-4 remains statistically insignificant even when controlling for the impact of operations, highlighting that both models perform similarly on this task. 
\begin{table}[t]
\centering
\resizebox{\linewidth}{!}{
\begin{tabular}{lcccc}
\toprule
\textbf{Predictor}& \textbf{Coefficient}& \textbf{SE}& \textbf{Z}&\textbf{p}\\
\midrule
 Constant& 1.648& 0.182& 9.053&0.000**\\
 GPT-4 Turbo& -0.053& 0.251& -0.212&0.832\\
 GPT-3.5 Turbo& -0.735& 0.235& -3.121&0.002**\\
 LLEMMA& -2.441& 0.267& -9.138&0.000**\\
 MAmmoTH& -1.009&0.231& -4.363&0.000**\\
Llama-2& -0.587& 0.241&  -2.435&0.015*\\
\midrule
 % & & & & \\
 Constant& 1.496& 0.254& 5.895&0.000**\\
 GPT-4 Turbo& 0.064& 0.262& 0.243&0.808\\
 GPT-3.5 Turbo& 	-0.496& 0.247& -2.008&0.045*\\
 LLEMMA& -2.279& 0.276& -8.269& 0.000**\\
 MAmmoTH& -0.867&0.238&-3.638& 	0.000**\\
 Llama-2&-0.532& 0.244&-2.183& 0.029*\\
 Addition&0.130& 0.153& 0.850& 0.395\\
 Subtraction& 0.234& 0.165& 1.413& 0.158\\
 Multiplication& -0.123& 0.165& -0.748& 0.454\\
 Division&-0.124& 0.193& -0.645& 0.519\\
 Fractions&-0.152& 0.323&-0.471& 0.638\\
 Decimals& -0.345& 0.208&-1.658& 0.097\\
 \bottomrule
\end{tabular}}
\caption{\label{citation-guide}Logistic regression results for meets all criteria (MaC), with and without controlling for the impact of question type. These results only consider questions which are labeled as solvable. The reference model for the constant is MATHWELL. A * or ** indicates statistical significance at the p<0.05 or p<0.01 level, respectively.
}
\label{mac_regression}
\end{table}
\subsection{Accuracy by Question Type}
As shown in Table \ref{accuracy_by_type}, MATHWELL's accuracy does not differ significantly or substantively by operation, as its accuracy remains above 90\% for all operations, while the other models except GPT-4 have a significant and/or substantive gap in their accuracy for the operation they perform best on relative to the operation they perform worst on. While MAmmoTH outperforms MATHWELL for addition, multiplication, and division, MATHWELL performs better in the other three operations and in overall accuracy. Further, MATHWELL outperforms GPT-4 in addition, subtraction, and decimals while maintaining similar performance in the other three operations. 
\begin{table}[t]
\centering
\resizebox{\linewidth}{!}{
\begin{tabular}{clcccccc}
\toprule
&\textbf{Model}& \textbf{Add.}& \textbf{Sub.}& \textbf{Mult.}&\textbf{Div.}  & \textbf{Frac.}&\textbf{Dec.}\\
\midrule
 \parbox[t]{2mm}{\multirow{2}{*}{\rotatebox[origin=c]{90}{API}}} & 
GPT-4 Turbo& 96.0& 94.3& 95.9& 93.3& 100.0&90.9\\
& \textbf{GPT-3.5 Turbo}& 91.0& 95.2& 86.6& 87.5& 71.4&90.2\\
 \midrule
\parbox[t]{2mm}{\multirow{4}{*}{\rotatebox[origin=c]{90}{{Public}}}} &
LLEMMA& 76.2& 72.3& 63.4& 56.0& 50.0&63.2\\
& MAmmoTH& \textbf{96.5}& 96.3& \textbf{96.8}&\textbf{97.6}& 87.5&91.3\\
& \textbf{Llama-2}& 89.3& 91.1& 87.5&  80.0& 82.4&75.0\\
& MATHWELL& 96.1& \textbf{97.4}&  94.5&91.3& \textbf{100.0}&\textbf{94.1}\\
\bottomrule
\end{tabular}}
\caption{\label{citation-guide}Accuracy by operation. Add., Sub., Mult., Div., Frac., Dec., No Ops, Total Ops, and MaC are addition, subtraction, multiplication, division, fractions, decimals, no operations, total operations, and meets all criteria, respectively. Bold indicates the best open-source performance in each operation. A bold model name indicates the difference between that model's operation with the highest accuracy and lowest accuracy is statistically significant at the p<0.05 level. 
}
\label{accuracy_by_type}
\end{table}
\begin{figure}[t]
    \centering
    \includegraphics[width=1\linewidth]{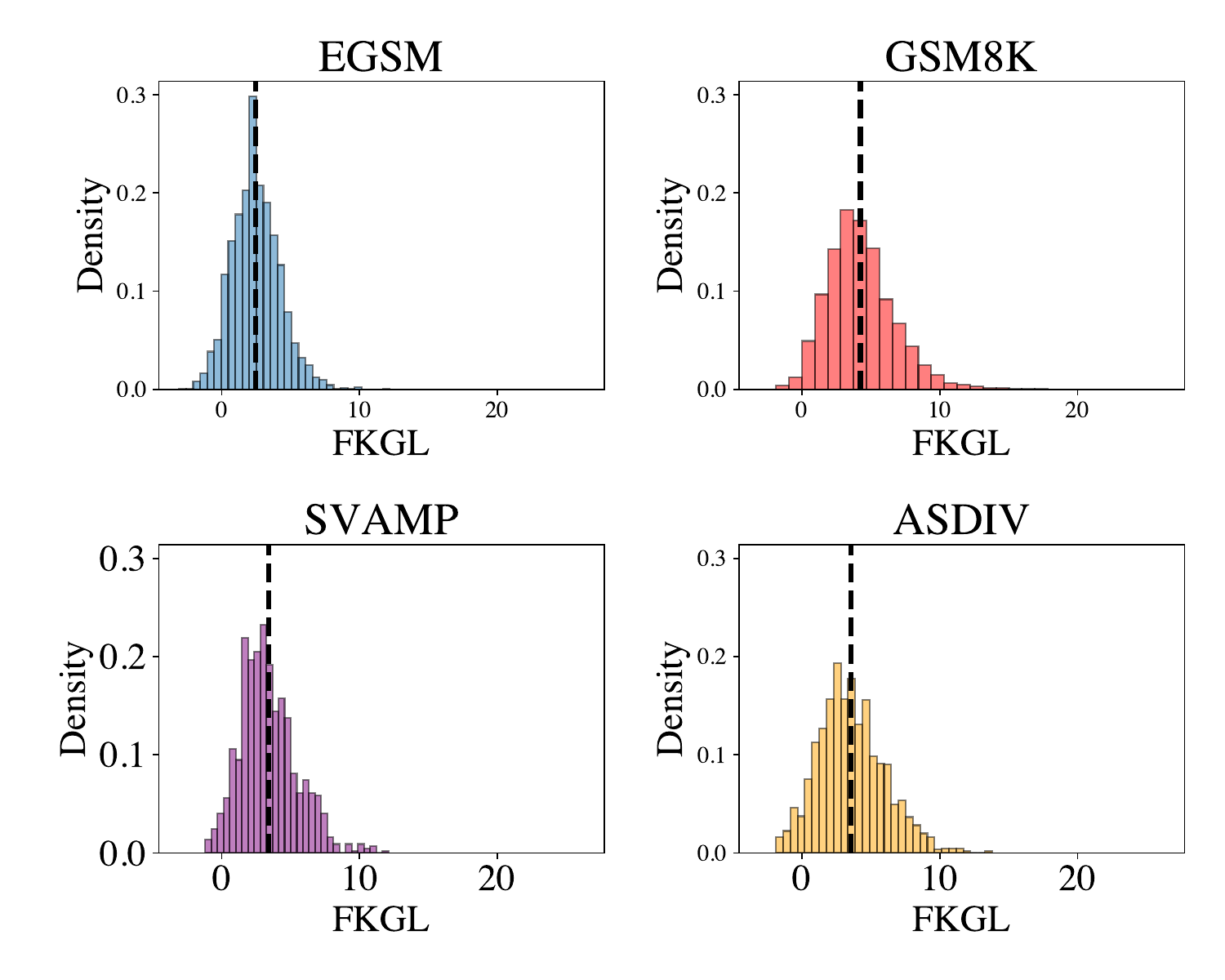}
    \caption{Flesch-Kincaid grade level (FKGL) distribution of training datasets. Dotted lines show the mean for each dataset.}
    \label{dataset_read}
\end{figure}

\subsection{FKGL Comparisons}
\begin{figure}[t]
    \centering
    \includegraphics[width=1\linewidth]{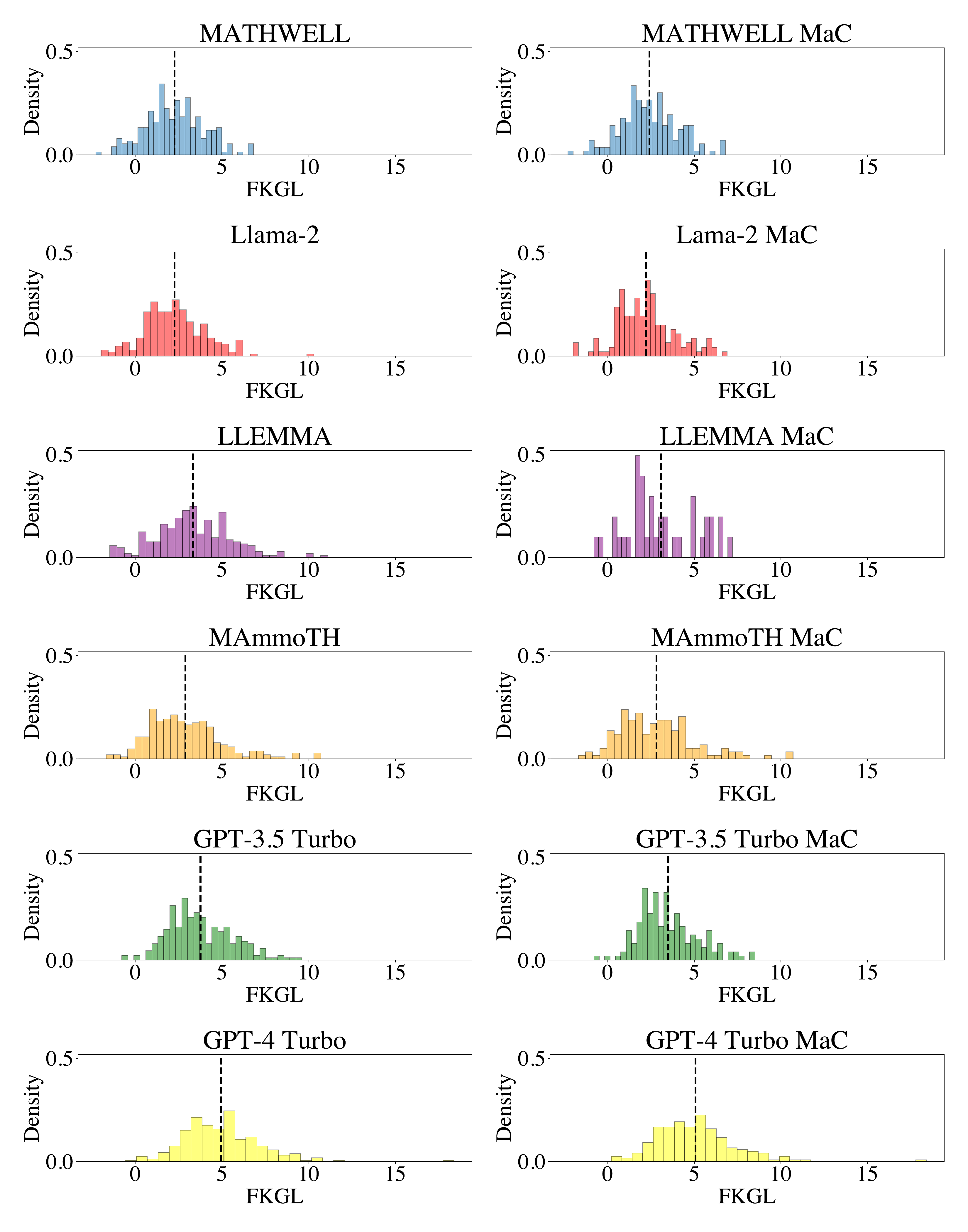}
    \caption{Flesch-Kincaid grade level (FKGL) distribution of model generations for all versus MaC questions. Dotted lines show the mean for each model.}
    \label{fig:model_read_comp}
\end{figure}

As shown in Figure \ref{fig:model_read_comp}, the FKGL distribution for all versus MaC questions does not significantly vary for all models except for LLEMMA, whose MaC questions tend to have lower FKGL than its average question. Figure \ref{dataset_read} compares the FKGL distribution of EGSM questions to the three other existing datasets that are mathematically appropriate for grade school students. Fewer of EGSM's questions than other existing datasets are written at a grade level beyond 8th grade.  
\subsection{Additional Automatic Evaluation Comparisons}
As shown in Table \ref{tab:additional_auto_eval_models}, the findings from the PPL experiment remain unchanged when PPL is evaluated with GPT-2 (MATHWELL outputs still have the lowest PPL of all models considered). Additionally, PPL does not significantly vary for all models except for LLEMMA when comparing all to MaC generations. In the table, we also compare the BERTScore between all and MaC questions from each model and show they are similar. While LLEMMA has a longer average token length for all its generations than MATHWELL, MATHWELL's MaC average token length is the longest of the open-source models considered. 
\begin{table*}
\centering
\resizebox{\linewidth}{!}{
\begin{tabular}{clccccccccc}
\toprule
& \textbf{Model}&  \textbf{GPT-2 PPL $\downarrow$}&\textbf{PPL $\downarrow$} & \textbf{MaC PPL $\downarrow$}& \textbf{BF1}& \textbf{MaC BF1}& \textbf{All/MaC BF1}&\textbf{GSM BF1}& \textbf{Length}&\textbf{MaC AL}\\
\midrule
\parbox[t]{2mm}{\multirow{2}{*}{\rotatebox[origin=c]{90}{API}}} &
GPT-4 Turbo&   10.68 (2.90) &2.50 (0.03)& 2.49 (0.03)& 85.4& 85.5& 85.5& 84.6& \textbf{66.0 (2.12)}& 66.8 (2.62)\\
& GPT-3.5 Turbo&   10.57 (3.60)  &2.64 (0.03)& 2.70 (0.04)& 85.6& 85.8& 85.7& 84.6& 52.8 (1.00)& 49.9 (1.16)\\
\midrule
\parbox[t]{2mm}{\multirow{4}{*}{\rotatebox[origin=c]{90}{Public}}} &
LLEMMA&   15.20 (8.40)  &3.82 (0.10)& 3.14 (0.10)& 84.3& 85.3& 84.6& 84.0& \textit{56.4 (1.44)}& 50.9 (2.89)\\
&MAmmoTH&   11.96 (4.26)  &2.76 (0.03)& 2.74 (0.04)& 86.0& 86.4& 86.1&84.6& 45.9 (1.13)&44.4 (1.15)\\
& Llama-2&   9.97 (4.12)   &2.52 (0.03)& 2.52 (0.04)& 85.5& 85.8& 85.6&84.3& 51.6 (0.99)&49.8 (1.19)\\
& MATHWELL (Ours)&   \textbf{9.90 (3.84)}&\textbf{2.44 (0.03)}& \textbf{2.43 (0.03)}& 85.5& 85.7& 85.6&84.2& 54.7 (0.86)&\textbf{54.1 (0.97)}\\
\bottomrule
\end{tabular}
}
\caption{\label{citation-guide}Automatic evaluation metrics for each model. PPL is perplexity (evaluated by Llama-2 70B unless otherwise noted), BF1 is BERTScore F1, MaC is meets all criteria (MaC), All/MaC BF1 compares all to MaC questions, GSM BF1 compares each model's questions to GSM8K, and Length is average token length. Bold indicates the lowest PPL and longest length for open-source models. Standard errors, where applicable, are in parentheses.
}
\label{tab:additional_auto_eval_models}
\end{table*}
\subsection{Exploring Automatic Classification} \label{sec:classifier}
\begin{table*}
\centering
\resizebox{\linewidth}{!}{
\begin{tabular}{lcccccccccccccccccc}
\toprule
& \multicolumn{6}{c}{\textbf{Solvability}} & \multicolumn{6}{c}{\textbf{Accuracy}}&\multicolumn{6}{c}{\textbf{Appropriateness}}\\
\cmidrule(lr){2-7} \cmidrule(lr){8-13} \cmidrule(lr){14-19}
 \textbf{Source}& \textbf{Acc.}& \textbf{BA}& \textbf{P}& \textbf{R}& \textbf{F1}&\textbf{RA}& \textbf{Acc.}& \textbf{BA}& \textbf{P}& \textbf{R}& \textbf{F1}& \textbf{RA}& \textbf{Acc.}& \textbf{BA}& \textbf{P}& \textbf{R}& \textbf{F1}&\textbf{RA}\\
\midrule
 Annotated Test& 78.9& 61.9& 86.7& 88.0& 87.4&0.701& 81.6& 53.8& 87.2& 92.2& 89.6& 0.649& 82.7& 72.0& 89.8& 88.9& 89.4&0.793\\
 GPT-4 Turbo& 83.6& 62.3& 96.2& 86.1& 90.9& 0.560& 95.8& 54.8& 96.2& 99.6& 97.8& 0.637& 77.2& 59.0& 87.2& 85.5& 86.4&0.579\\
 GPT-3.5 Turbo& 81.9& 52.3& 88.5& 91.3& 89.9& 0.519& 88.6& 55.2& 90.5& 97.4& 93.9& 0.541& 78.1& 64.3& 82.1& 91.0& 86.3&0.696\\
LLEMMA & 48.8&  49.8&  48.7& 90.2&63.2&0.542& 59.0& 47.1& 62.5& 89.7& 73.7& 0.492& 45.1& 50.1& 41.8& 80.4& 55.0&0.489\\
MAmmoTH & 84.4&  52.5&  87.4& 95.9&91.4&0.700& 93.5& 57.9& 95.7& 97.6& 96.6& 0.669& 66.8& 54.2& 69.8& 89.8& 78.6&0.574\\
 Llama-2 & 82.4&  53.1&  84.9& 96.2&90.2&0.619& 85.2& 51.6& 89.8& 94.1& 91.9& 0.677& 73.8& 50.4& 81.1& 88.2& 84.5&0.628\\
MATHWELL & 85.2&  54.3&  90.1& 93.7&91.9&0.599& 94.6& 62.7& 97.7& 96.8& 97.2& 0.790& 77.6& 63.1& 90.4& 82.9& 86.5&0.669\\
\bottomrule
\end{tabular}
}\caption{\label{citation-guide}DistilBERT \cite{DBLP:journals/corr/abs-1910-01108} text classifier performance for solvability, accuracy, and appropriateness. Acc., BA, P, R, and RA are accuracy, balanced accuracy, precision, recall and ROC AUC, respectively. 
}
\label{classifier}
\end{table*} 

Given the high cost of human evaluation, we consider automatically evaluating model outputs by training DistilBERT \cite{DBLP:journals/corr/abs-1910-01108} classifiers for solvability, accuracy and appropriateness using our annotated data, excluding the evaluation samples from each model. We use an 80/10/10 train, validation and test split. The solvability dataset has 3,234 rows, the accuracy dataset has 2,830 rows, and the appropriateness dataset has 2,660 rows. We exclude unsolvable questions from the accuracy and appropriateness datasets to promote specialization, as these questions do not have a solution and are inappropriate. Each dataset is unbalanced, with 82.3\% of questions labeled as solvable, 85.8\% labeled as accurate, and 80.1\% labeled as appropriate. As a result, we modify the training objective to weight the loss based on the inverse proportion of observations from each class. We train each model for 8 epochs using the HuggingFace library \cite{wolf_huggingfaces_2020} with strict regularization (weight decay = 0.9) and use the checkpoint with the lowest validation loss for testing.
\newline \indent As shown in Table \ref{classifier}, despite weighting the loss, each classifier has a lower balanced accuracy than its overall accuracy. The classifiers perform similarly on outputs from all models except LLEMMA, for which performance is significantly lower. As would be expected, performance for all models tends to be lower than on the test dataset. Excluding LLEMMA and the appropriateness classifier's precision on MAmmoTH outputs, each classifier has fairly high precision, recall and F1, suggesting they are effectively able to label solvable, accurate, and appropriate questions. The classifiers tend to have low ROC AUC across sources, suggesting they do not perform well at different prediction cutoffs. As a whole, Table \ref{classifier} provides evidence that text classifiers can learn some features that are important for labelling MATHWELL outputs, but future research should use more balanced datasets to improve their ability to label unsolvable, inaccurate, and inappropriate questions. 
%While the accuracy classifier has high precision, it struggles with recall, as the modified training objective prefers false negatives to positives---providing inaccurate solutions is worse than filtering out accurately-solved questions. 
\subsection{Important Criteria for Reference-free Word Problem Generator Training Data} \label{important_criteria}
In addition to high-quality grammar and problems that are similar to those students encounter in the classroom, the characteristics we find most important for training reference-free word problem generators are PoT solutions written as Python functions and questions that are educationally appropriate for K-8 students. Regarding the former, when we modified our prompt to ask for a Python function solution instead of a Python code solution, the share of question/answer pairs with executable code from an early version of MATHWELL increased from 18.9\% to 29.0\%. For educational appropriateness, when we used the GSM-Hard \cite{gao_pal_2023} dataset as part of MATHWELL's training data, the model often generated questions with large numbers that are inappropriate for K-8 students. We further show that existing data are not educationally appropriate in Section \ref{existing_datasets}. As shown in Table \ref{dataset_comparison}, EGSM is the only math QA dataset that has these two characteristics.
\subsection{Early MATHWELL Experimentation}\label{sec:experimentation}
In addition to training reference-free question/answer pair generators, we also experiment with training reference-free question generation models. Our theory is that if we could train a model to generate questions effectively, we could pass those questions to a math QA model to retrieve answers automatically. To test this theory, we finetune both Llama-2 and MAmmoTH as question generators using the same training data discussed in Appendix \ref{qlora},  except for excluding the solution for each question and modifying the standard prompt to ask the model to generate a question only. We then sample and evaluate 100 generations from each model. We find that MAmmoTH performs better than Llama-2 at this task, but neither model performs optimally. For example, only 19\% of the MAmmoTH generations include the requested topic and 52.6\% are solvable. Therefore, based on the results we report in Table \ref{main_results}, we conclude that it is more efficient to train a reference-free question/answer pair generator than question generator. 

\section{Finetuning Details} \label{finetuning_details}
\subsection{Initial Finetuning}  \label{qlora}
We re-format MathInstruct GSM8K \cite{yue_mammoth_2023} into an Alpaca-style \cite{alpaca} instruction dataset of question/answer pairs with a standard prompt asking the model to generate a grade school math word problem. We also include MathInstruct MATH and TheoremQA PoT \cite{yue_mammoth_2023} question/answer pairs with a standard prompt asking the model to generate a challenge math problem to expose the model to more complex code and further promote mathematical reasoning. We then finetune Llama-2 (70B) on this dataset of 25,926 rows using QLoRA \cite{dettmers_qlora_2023} for 4,250 steps. \textbf{Our finetuning process uses the following training resources and hyperparameters:} two A100 GPUs, a learning rate of 1e-4, LoRA modules at every model layer, per-device batch size of 1, and a 3\% warm-up ratio. The training script in the GitHub repo linked to this paper contains a full list of hyperparameters. 
%using the HuggingFace transformers library \cite{wolf_huggingfaces_2020} and bitsandbytes for quantization \cite{dettmers_bitsandbytes_2024}. 
% \subsubsection{Synthetic Data Generation and Further Finetuning} \label{data_generation}
% \noindent\textbf{Synthetic Data Generation and Further Finetuning.}
\subsection{Secondary Finetuning on Expert-annotated Data} \label{data_generation}
%Inspired by recent works that iteratively finetune LLMs \cite{zheng_gpt-fathom_2023}, 
Inspired by recent works that iteratively finetune LLMs \cite{guo2024humaninstructionfree, wang2024steponfeet}, including OpenAI's FeedME approach \cite{zheng_gpt-fathom_2023}, we next aim to evaluate and improve how well MATHWELL meets our three main criteria: solvability, accuracy, and appropriateness.
However, these properties are unannotated in prior datasets and we find they are ill-suited for training educational generators in Section \ref{existing_datasets}.
Therefore, we few-shot prompt the finetuned model and generate 3,234 problems.
% Because there is no existing age-appropriate math QA dataset with Python function solutions,
% % which we find important for training reference-free word problem generators (see Section \ref{sgsm_characteristics}),
% we few-shot prompt our finetuned model to generate synthetic data with Python function solutions, resulting in 3,234 problems.
% Using Zooniverse \cite{noauthor_zooniverse_nodate}
We then acquire annotations for solvability, accuracy, and appropriateness from teachers. Further details about the annotation process and annotator agreement are in Section \ref{EGSM_datasets} and Appendix \ref{sec:annotation}. These annotations identified 1,906 problems that are simultaneously solvable, accurate, and appropriate, or meet all criteria (MaC).
We then finetune MATHWELL on these MaC outputs for 1,250 more steps using the same hyperameters discussed above.
% This multi-iteration training procedure akin to FeedME approach OpenAI used to train text-davinci-002 \cite{zheng_gpt-fathom_2023}. 

\section{Annotated Example Generations}\label{sec:examples}
The sections below display randomly sampled generations from each model, including examples of both questions that do not meet evaluation criteria and those that do. For the examples that do not meet the specified criteria, we provide a short natural language annotation explaining the error. 

\subsection{GPT-4 Turbo}
\subsubsection{Solvability}
\paragraph{Positive Example} In a volleyball match, Team A scored twice as many points as Team B. If Team B scored 15 points, how many points in total were scored in the match?

\paragraph{Negative Example} In a Fortnite Battle Royale match, there are 100 players. If one player can eliminate 4 opponents, how many players in the game can achieve such an elimination count, assuming no other players are eliminated?

\subparagraph{Negative Example Reasoning} It is not entirely clear what this question is asking. If it defined how many players should be left in the game after eliminations, it would be possible to determine how many players can be eliminated but, without defining that, it is not clear what the student needs to do to solve this problem. 

\subsubsection{Accuracy}
\paragraph{Positive Example}
Jenny has 15 cats. She buys 2 new cat towers, each can hold 4 cats. If she places as many cats as possible on the cat towers, how many cats are not on the towers?
\newline \newline
Solution: 
\newline def solution(): \newline
\indent \# Jenny starts with 15 cats \newline
\indent total\_cats = 15 \newline
\indent \# Each new cat tower holds 4 cats \newline
\indent tower\_capacity = 4 \newline 
\indent \# She buys 2 new cat towers \newline 
\indent total\_towers = 2 \newline
\indent \# The total number of cats the towers can hold is \newline
\indent total\_capacity = tower\_capacity * total\_towers \newline
\indent \# The number of cats not on the towers is \newline
\indent cats\_not\_on\_towers = total\_cats - total\_capacity \newline
\indent return cats\_not\_on\_towers \newline \newline
\# Uncomment the line below to test the function with the given values \newline 
\# print(solution())\footnote{As discussed in Appendix \ref{sec:gpt4_coding}, GPT-4 tends to output additional information beyond what is requested in the prompt and shown in few-shot examples, such as example usage or comments about the Python function solution, as is included here.}

\paragraph{Negative Example} \label{gpt4_extra}
Mrs. Johnson has 12 adult cats and each adult cat has 3 kittens. After several months, those kittens grow up and each new adult cat also has 2 kittens. How many kittens are there in total now?
\newline \newline
Solution: \newline
def solution(): \newline 
\indent \# Mrs. Johnson has 12 adult cats originally\newline 
\indent adult\_cats\_initial = 12 \newline
\indent \# Each adult cat has 3 kittens\newline
\indent kittens\_per\_adult\_initial = 3\newline 
\indent \# Calculate the initial number of kittens \newline
\indent initial\_kittens = adult\_cats\_initial * \newline
\indent \indent kittens\_per\_adult\_initial\newline
\indent \# The initial kittens grow up to become adults \newline
\indent new\_adult\_cats = initial\_kittens\newline
\indent  \# Each new adult cat has 2 kittens\newline
\indent kittens\_per\_new\_adult = 2 \newline
\indent \# Calculate the new kittens born from the \newline
\indent \# grown-up kittens \newline
\indent new\_kittens = new\_adult\_cats * \newline
\indent \indent kittens\_per\_new\_adult \newline
\indent \# The total number of kittens now is the sum \newline
\indent \# of the initial kittens and new kittens \newline
\indent total\_kittens = initial\_kittens + new\_kittens \newline
\indent return total\_kittens\newline \newline
\# Call the function and print the result \newline
print(f"There are \{solution()\} kittens in total now.")

\subparagraph{Negative Example Reasoning} The solution adds the original number of kittens to the number of new kittens, which is incorrect because the original kittens are now considered adults based on the text of the question.

\subsubsection{Appropriateness}
\paragraph{Positive Example} During a week of intense battles, the Power Rangers defeated 15 evil monsters on Monday, twice as many on Tuesday as on Monday, and 10 fewer on Wednesday than on Tuesday. How many monsters did they defeat in total from Monday to Wednesday?

\paragraph{Negative Example} Serena Williams has won a certain number of tennis matches this season. If she wins 8 more matches, she will double the number she has won so far. How many matches has Serena won this season?
    
\subparagraph{Negative Example Reasoning} This question does not require any mathematical operation to solve and is therefore not appropriate. The question tells you Serena's number of wins will double if she wins 8 more matches, which means she had to have won 8 matches so far.

\subsection{GPT-3.5 Turbo}
\subsubsection{Solvability}
\paragraph{Positive Example} During a basketball game, the Lakers made a total of 84 points. LeBron James scored 35 points, and Anthony Davis scored 20 points. How many points were scored by the rest of the Lakers team?

\paragraph{Negative Example} The Ninja Turtles went to the pizza parlor and ordered 8 pizzas. Each pizza had 8 slices. If they each ate 2 slices, how many slices of pizza were left over?

\subparagraph{Negative Example Reasoning} This question does not define the number of Ninja Turtles. While many students might know there are typically 4 main Ninja Turtles, not every student would know this and all math questions should still define the key variables you need to answer them. 

\subsubsection{Accuracy}
\paragraph{Positive Example}
Black Panther has 1500 vibranium shards. He needs to divide them equally amongst his 5 closest allies. How many vibranium shards will each ally receive?
\newline \newline
Solution: 
\newline def solution(): \newline
\indent \#Black Panther has 1500 vibranium shards \newline
\indent vibranium\_total = 1500 \newline
\indent\#He needs to divide them equally amongst his 5 \newline
\indent \#allies \newline
\indent allies = 5 \newline 
\indent \#The answer is \newline 
\indent result = vibranium\_total / allies \newline
\indent return result 

\paragraph{Negative Example}
A dog shelter has 40 dogs. If each dog needs 2 cups of food per day, and a bag of food contains 40 cups, how many bags of food will the shelter need for a 30-day period?
\newline \newline
Solution: \newline
def solution(): \newline 
\indent\#The shelter has 40 dogs\newline 
\indent dogs = 40 \newline
\indent \#Each dog needs 2 cups of food per day\newline
\indent food\_per\_dog = 2\newline 
\indent \#A bag of food contains 40 cups\newline
\indent food\_per\_bag = 40\newline
\indent \#For 30 days, the shelter will need\newline
\indent total\_food\_needed = dogs * food\_per\_dog * 30 \newline
\indent \#The answer is\newline
\indent result = total\_food\_needed // food\_per\_bag + 1 \newline
\indent return result 

\subparagraph{Negative Example Reasoning} The solution adds an additional bag of food to the total assuming that the answer is a decimal and, therefore, that the shelter would need to buy another full bag since a partial bag is not possible. However, since the answer is whole number, this additional bag leads to outputting an incorrect result. 

\subsubsection{Appropriateness} \label{too_hard_example}
\paragraph{Positive Example} A cheerleading team has 20 members. They want to evenly divide into 4 squads for a competition. How many members will be on each squad?

\paragraph{Negative Example} In a Pokémon battle, Pikachu has a 60\% chance of winning each round. If Pikachu and Ash battle for 5 rounds, what is the probability that Pikachu wins all 5 rounds?

\subparagraph{Negative Example Reasoning} This question is too hard for a middle school student. It assumes a student knows about the probability of independent events, which is typically not covered until high school or an introductory statistics college course. 

\subsection{LLEMMA}
\subsubsection{Solvability}
\paragraph{Positive Example}
Steph Curry makes 2133 free throws for the year. He hits 342 more free throws than he misses. How many free throws does Steph Curry miss?
\paragraph{Negative Example}
There are 5175 Pokémon available to battle in Pokémon Sword and Shield. There are 310 ground-type Pokémon and 182 water-type Pokémon. There are 610 Pokémon that are not ground nor water type. How many Pokémon are ground-type or water-type?
\subparagraph{Negative Example Reasoning} The question contains conflicting information in that it defines the number of ground and water type Pokémon twice.

\subsubsection{Accuracy}
\paragraph{Positive Example}
Naruto eats 4 bananas a day. How many bananas does he eat in a month if there are 30 days in a month? 
\newline \newline
Solution: 
\newline def solution(): \newline
\indent \#Naruto eats 4 bananas a day \newline
\indent bananas\_per\_day = 4 \newline
\indent \#There are 30 days in a month \newline
\indent days\_in\_month = 30 \newline 
\indent \#The answer is \newline 
\indent result = bananas\_per\_day * days\_in\_month \newline
\indent return result

\paragraph{Negative Example}
12 cats eat 400 pounds of cat food every Saturday morning. 15 cats eat 500 pounds of cat food on Sunday morning. How many pounds of cat food are eaten in total?
\newline \newline
Solution: \newline
def solution(): \newline 
\indent \#12 cats eat 400 pounds of cat food every \newline 
\indent \#Saturday morning \newline
\indent cat\_food\_satuday\_morning = 12*400 \newline
\indent \#15 cats eat 500 pounds of cat food on Sunday \newline \indent \#morning \newline
\indent cat\_food\_sunday\_morning = 15*500 \newline
\indent \#In total, \newline
\indent result = cat\_food\_satuday\_morning + \newline \indent \indent cat\_food\_sunday\_morning \newline
\indent return result
    
\subparagraph{Negative Example Reasoning} The model multiplies the number of cats by the amount of cat food eaten, rather than adding the two sums of cat food together. 

\subsubsection{Appropriateness} \label{inappropriate_content}
\paragraph{Positive Example} Hello Kitty makes 18 bracelets in 4 hours. How many bracelets per hour does she make?

\paragraph{Negative Example} Batman caught a baddie with his trademark punch. Each punch knocks out 7 baddies. If Batman has thrown 60 punches, how many baddies has he knocked out?
    
\subparagraph{Negative Example Reasoning} While this question is comical, it is not appropriate for a K-8 student because it involves physically harming another person. 

\subsection{MAmmoTH}
\subsubsection{Solvability} 
\paragraph{Positive Example} The 5 Power Rangers, Red, Blue, Black, Yellow, and Pink, are eating pizza. Red ate 31 slices, Blue ate 28 slices, Black ate 23 slices, Yellow ate 21 slices, and Pink ate 15 slices. How many slices of pizza did they eat in total?

\paragraph{Negative Example} In Fortnite, the player has 5 health points. When the player is hit by an enemy, they lose 2 health points. How many health points does the player have left?

\subparagraph{Negative Example Reasoning} This question defines how many health points a player loses when they get hit by an enemy, but it does not define how many times the player got hit, so it is impossible to determine how many health points they have left. 

\subsubsection{Accuracy}
\paragraph{Positive Example} Harry Potter and his friends have just finished their exams and are looking forward to a well-deserved break. They decide to go on a camping trip together. They have 120 Galleons between them. They spend 30 Galleons on food in the morning and 20 Galleons on food in the afternoon. They have 20 Galleons left. How many Galleons did they spend in the evening? 
\newline \newline 
Solution: \newline
def solution(): \newline
\indent \# total amount of Galleons they had \newline
\indent total\_galleons = 120 \newline
\indent \# amount spent on food in the morning \newline 
\indent morning\_food = 30 \newline
\indent \# amount spent on food in the afternoon \newline
\indent afternoon\_food = 20 \newline
\indent \# amount left \newline
\indent left\_galleons = 20 \newline
\indent \# calculate the amount spent in the evening \newline
\indent evening\_food = total\_galleons - morning\_food - \newline 
\indent \indent afternoon\_food - left\_galleons \newline
\indent return evening\_food \newline
print(solution())
\footnote{MAmmoTH is prone to ending its generations with print(solution()) because its PoT training data always ends with a print statement \cite{yue_mammoth_2023}.}

\paragraph{Negative Example} Captain Marvel has 100 friends on Facebook. She has 40 more friends than the average number of friends her friends have. How many friends does the average friend of Captain Marvel have? 
\newline \newline 
Solution: \newline
def solution(): \newline
\indent \#Captain Marvel has 100 friends on Facebook \newline
\indent captain\_marvel\_friends = 100 \newline
\indent \#She has 40 more friends than the average \newline 
\indent \#number of friends her friends have \newline 
\indent captain\_marvel\_friends\_more = 40 \newline 
\indent \#The average number of friends her friends have \newline 
\indent average\_friends = (captain\_marvel\_friends - \newline 
\indent \indent captain\_marvel\_friends\_more) / 2 \newline 
\indent return average\_friends

\subparagraph{Negative Example Reasoning} The model calculates an average rather than subtract the average number of friends Captain Marvel's friends have from her number of friends.

\subsubsection{Appropriateness} \label{strange_question}
\paragraph{Positive Example} Taylor Swift has 11 Grammys, 29 AMAs, 12 CMAs, 8 ACMs and 35 BMAs. How many awards has she won in total?

\paragraph{Negative Example} The soccer team has 32 players. Each player has 2 legs. How many legs does the team have?

\subparagraph{Negative Example Reasoning} While this question is solvable, it is not appropriate because it is strange to ask how many legs a soccer team has. 

\subsection{Llama-2}
\subsubsection{Solvability}
\paragraph{Positive Example} The Pittsburgh Steelers have 11 players on the field. 3 of the players are from the 10th grade, 6 are from the 11th grade, and the rest are from the 12th grade. How many players are from the 12th grade?

\paragraph{Negative Example} The cheerleaders at the school are planning a pep rally. They need to make 200 pom-poms. Each pom-pom requires 20 minutes to make. How many hours will it take to make 200 pom-poms?

\subparagraph{Negative Example Reasoning} The question does not define how many cheerleaders there are, so it is impossible to determine how long it will take them to make the pom-poms. 

\subsubsection{Accuracy}
\paragraph{Positive Example} Barbie has 100 pairs of shoes. She gave 20 pairs of shoes to Ken, 30 pairs of shoes to Skipper, and 40 pairs of shoes to Stacie. How many pairs of shoes does Barbie have left? 
\newline \newline
Solution: \newline
def solution(): \newline
\indent \#Barbie started with 100 pairs of shoes \newline
\indent shoes\_initial = 100 \newline
\indent \#She gave 20 pairs of shoes to Ken \newline
\indent shoes\_ken = 20 \newline
\indent \#She gave 30 pairs of shoes to Skipper \newline 
\indent shoes\_skipper = 30 \newline
\indent \#She gave 40 pairs of shoes to Stacie \newline
\indent shoes\_stacie = 40 \newline
\indent \#The answer is\newline
\indent result = shoes\_initial - shoes\_ken - \newline \indent \indent shoes\_skipper - shoes\_stacie \newline
\indent return result

\paragraph{Negative Example} A basketball team scored 120 points in a game. The team scored 30 points in the first quarter, 35 points in the second quarter, 20 points in the third quarter, and 35 points in the fourth quarter. How many points did the team score in the second half of the game? \newline \newline 
Solution: \newline 
def solution(): \newline 
\indent \#The team scored 120 points in a game \newline 
\indent total\_points = 120 \newline
\indent \#30 points in the first quarter \newline 
\indent points\_first\_quarter = 30 \newline 
\indent \#35 points in the second quarter \newline
\indent points\_second\_quarter = 35 \newline
\indent \#20 points in the third quarter \newline
\indent points\_third\_quarter = 20 \newline
\indent \#35 points in the fourth quarter \newline
\indent points\_fourth\_quarter = 35 \newline 
\indent \#The answer is \newline 
\indent result = points\_second\_quarter + \newline \indent \indent points\_third\_quarter + points\_fourth\_quarter \newline
\indent return result

\subparagraph{Negative Example Reasoning} The model correctly defines the variables, but it incorrectly adds the points scored in the second quarter to the total for the points scored in the second half. 

\subsubsection{Appropriateness} \label{unrealistic_question}
\paragraph{Positive Example} Patrick Mahomes is a professional football player. He is 6 feet and 3 inches tall. 1 inch is 2.54 centimeters. How many centimeters tall is Patrick Mahomes?

\paragraph{Negative Example} A cat has 100 kittens. 20 of them are calico, 30 are tabby, and the rest are siamese. How many kittens are siamese?

\subparagraph{Negative Example Reasoning} This question is not based in reality, as it is not possible for one cat to birth 100 kittens, nor is it possible for them to be different breeds. 

\subsection{MATHWELL}
\subsubsection{Solvability}
\paragraph{Positive Example} A soccer team has 11 players on the field at a time. They have 22 more players on the bench. How many players are on the team in total?

\paragraph{Negative Example} Steph Curry, the NBA star, is 6.2 feet tall. He wears shoes that are 1.5 inches thick. How tall is he in inches? (1 foot is 12 inches)

\subparagraph{Negative Example Reasoning} This question is not solvable because it is vague: we do not know if we are calculating his height in inches with or without shoes on. 

\subsubsection{Accuracy}
\paragraph{Positive Example} In the game, Super Mario can stomp on 30 Goombas, 20 Koopas, and 10 Piranha Plants. How many enemies does he stomp on total?
\newline 
\newline 
Solution: \newline
def solution():\newline
\indent \#Super Mario can stomp on 30 Goombas \newline
\indent goombas = 30 \newline
\indent \#20 Koopas \newline
\indent koopas = 20 \newline
\indent \#and 10 Piranha Plants \newline
\indent piranha\_plants = 10 \newline 
\indent \#The answer is \newline 
\indent result = goombas + koopas + piranha\_plants \newline
\indent return result

\paragraph{Negative Example} A Minecraft player has 100000 blocks. 20000 of the blocks are dirt, 30000 of the blocks are stone, 20000 of the blocks are wood, and 30000 of the blocks are diamond. How many of the blocks are not dirt, stone, wood, or diamond? 
\newline 
\newline
Solution: \newline
def solution(): \newline
\indent \#100000 blocks \newline
\indent blocks\_total = 100000 \newline
\indent \#20000 of the blocks are dirt \newline 
\indent blocks\_dirt = 20000 \newline 
\indent \#30000 of the blocks are stone \newline
\indent blocks\_stone = 30000 \newline 
\indent \#20000 of the blocks are wood \newline
\indent blocks\_wood = 20000 \newline
\indent  \#30000 of the blocks are diamond \newline
\indent blocks\_diamond = 30000 \newline
\indent \#The answer is \newline
\indent result = blocks\_total - blocks\_dirt - blocks\_stone \newline 
\indent \indent - blocks\_wood \newline 
\indent return result 

\subparagraph{Negative Example Reasoning} The model correctly defines the necessary variables, but does not subtract the number of diamond blocks from the total number of blocks.

\subsubsection{Appropriateness} 
\paragraph{Positive Example} LeBron James has 12000 points. He is 4000 points away from the all-time scoring record. How many more points does he need to average per game for the next 20 games to break the record?

\paragraph{Negative Example} \label{no_operation_example}
A field hockey team has 11 players. 3 of them are forwards, 3 of them are midfielders, 3 of them are defenders, and 2 of them are goalies. How many forwards are there?

\subparagraph{Negative Example Reasoning} This question is inappropriate to give to a student because it does not require any mathematical operations to solve. It directly defines the number of forwards on the team. 

\section{Topics for Data Generation}\label{sec:generation_topics}
We collected topics and keywords in collaboration with our volunteer annotators with K-12 teaching experience. The topics and keywords span a wide array of student interests including sports, music, video games, movies/TV, animals, vehicles, and food. On top of being manually collected from experts, we believe the resultant list is largely inarguable. 

We used a list of 43 topics when comparing models for the experiments reported in Section \ref{experiments} to make sure the generations could be compared fairly and followed a similar contextual distribution. The full list of topics written in Python list format is below. 

\paragraph{Topics}: ['Superman', 'Batman', 'Wonder Woman', 'Barbie', 'Power Rangers', 'basketball', 'soccer', 'football', 'volleyball', 'field hockey', 'Fortnite', 'Spiderman', 'Iron Man', 'Captain America', 'Captain Marvel', 'Thor, the God of Thunder', 'Ninja Turtles', 'Black Panther', 'Taylor Swift', 'swimming',\
'Pokémon', 'Super Mario', 'Naruto', 'unicorns', 'Hello Kitty', 'Minecraft', 'lacrosse', 'cheer leading', 'LeBron James', 'Steph Curry', 'Patrick Mahomes',\
'Serena Williams', 'dogs', 'cats', 'dinosaurs', 'Harry Potter', 'cars', 'planes', 'trains', 'pizza', 'cookies', 'ice cream', 'candy']

\section{GPT-4 Coding Performance}\label{sec:gpt4_coding}
GPT-4 tends to output additional information beyond what is requested in the prompt and shown in few-shot examples, such as example usage or comments about the Python function solution (see Appendix \ref{gpt4_extra} for an example). This results in the model having a lower percentage of executable code than GPT-3.5 when its output is parsed by the same script we used to parse the output from all other models we evaluated. We used the same parsing script across models to evaluate them all the same way, rather than creating a customized script for GPT-4.

\end{document}